\documentclass{article}

\PassOptionsToPackage{numbers, compress}{natbib}

\usepackage[final]{neurips_2024}

\usepackage[utf8]{inputenc} 
\usepackage[T1]{fontenc}    
\usepackage[hidelinks]{hyperref}       
\usepackage{url}            
\usepackage{booktabs}       
\usepackage{amsfonts}       
\usepackage{nicefrac}       
\usepackage{microtype}      
\usepackage{xcolor}         

\usepackage{graphicx}
\usepackage{amsmath}
\usepackage{amssymb}
\usepackage{mathtools}
\usepackage{amsthm}
\usepackage{fontawesome5}
\usepackage{enumitem}
\setlist[itemize]{noitemsep, topsep=0pt}
\usepackage{wrapfig}
\usepackage{algorithm}
\usepackage{wrapfig}

\usepackage{cleveref}
\usepackage{minitoc}
\usepackage{wrapfig}
\usepackage{bm}

\usepackage{minitoc}
\usepackage{tocloft}
\usepackage{amsmath}
\usepackage{algorithm}
\usepackage{algpseudocode}
\newtheorem{proposition}{Proposition}

\title{SPO: Sequential Monte Carlo Policy Optimisation}

\author{%
  Matthew V Macfarlane \thanks{Work done during internship at InstaDeep} \\
  University of Amsterdam\\
  \texttt{m.v.macfarlane@uva.nl} \\
  \And
  Edan Toledo \\
  InstaDeep \\
  \AND
  Donal Byrne \\
  InstaDeep \\
  \And
  Paul Duckworth \\
  InstaDeep \\
  \AND
  Alexandre Laterre \\
  InstaDeep \\
}

\begin{document}
\maketitle

\doparttoc 
\faketableofcontents 


\begin{abstract}

Leveraging planning during learning and decision-making is central to the long-term development of intelligent agents. Recent works have successfully combined tree-based search methods and self-play learning mechanisms to this end. However, these methods typically face scaling challenges due to the sequential nature of their search. While practical engineering solutions can partly overcome this, they often result in a negative impact on performance. In this paper, we introduce SPO: Sequential Monte Carlo Policy Optimisation, a model-based reinforcement learning algorithm grounded within the Expectation Maximisation (EM) framework. We show that SPO provides robust policy improvement and efficient scaling properties. 
The sample-based search makes it directly applicable to both discrete and continuous action spaces without modifications. We demonstrate statistically significant improvements in performance relative to model-free and model-based baselines across both continuous and discrete environments. Furthermore, the parallel nature of SPO’s search enables effective utilisation of hardware accelerators, yielding favourable scaling laws.

\end{abstract}
\section{Introduction}

The integration of reinforcement learning (RL) and neural-guided planning methods has recently achieved considerable success. Such methods effectively leverage planning during training via iterative imitation learning~\citep{anthony2017thinking,schrittwieser2020mastering,danihelka2021policy,antonoglou2021planning}. Applying additional computation via planning generates improved policies which is then amortised \cite{amos2023tutorial} into a neural network policy. Using this policy as part of planning itself creates a powerful self-improvement cycle.  They have demonstrated state of the art performance in applications ranging from chess~\citep{silver2018general} and matrix multiplication~\citep{AlphaTensor2022}, to language modelling~\citep{pang2023language}. 
However, one of the most commonly-used search-based policy improvement operators, MCTS~\citep{coulom2006efficient}: 
\romannumeral1)~performs poorly on low budgets \citep{grill2020monte}, 
\romannumeral2)~is inherently sequential limiting its scalability~\citep{liu2018watch,segal2010scalability}, and 
\romannumeral3)~requires modifications to adapt to large or continuous action spaces~\citep{moerland2018a0c, hubert2021learning}. 
These limitations underscore the need for more scalable, efficient and generally applicable algorithms. 

In this work, we introduce SPO: Sequential Monte Carlo Policy Optimisation, a model-based RL algorithm that utilises scalable sampled-based Sequential Monte Carlo planning for policy improvement. We formalise SPO as an approximate policy iteration algorithm, and show that it is formally grounded within the Expectation Maximisation (EM) framework. 

SPO is unique by leveraging both breadth and depth of search in order to provide better estimates of the target distribution, derived via EM optimisation, compared to previous works. When viewed relative to popular algorithms it combines the breadth used in MPO \citep{abdolmaleki2018maximum} and the depth used in V-MPO \citep{song2019v}, while enforcing KL constraints on updates to ensure stable policy improvement. SPO achieves strong performance across a range of benchmarks, outperforming AlphaZero, a leading model-based expert iteration algorithm. We also benchmark against a leading Sequential Monte Carlo (SMC) algorithm from \citet{piche2018probabilistic} and show that leveraging posterior estimates from SMC for policy improvement is central to strong performance.

We demonstrate that \romannumeral 1) SPO outperforms baselines across both high-dimensional continuous control and challenging discrete planning tasks without algorithmic modifications, and \romannumeral 2) the sampling-based approach is inherently parallelisable, enabling the use of hardware accelerators to improve training speed. This provides significant advantages over MCTS-based methods \citep{coulom2006efficient}, whose sequential process can be particularly inefficient during training. 3) The explicit use of KL targeting leads to strong and stable policy improvement, reducing the need for a grid search over exploration hyperparameters in search. We find that SPO, empirically has strong scaling behaviours, with improved performance given additional search budget both during training and inference.

\section{Related Work}

There is an extensive history of prior works that frame control and RL as an inference problem~\citep{toussaint2006probabilistic,levine2018reinforcement}. Expectation Maximization~\citep{dayan1997using,furmston2010variational,neumann2011variational} is a popular approach to directly optimising the evidence lower bound (ELBO) of the probability of optimality of a given policy. 
A number of model free methods implement EM, including RWR~\citep{peters2007reinforcement}, REPS~\citep{peters2010relative}, MPO~\citep{abdolmaleki2018maximum}, V-MPO~\citep{song2019v}, and AWAC~\citep{nair2020awac} each with different approaches to the E-step and M-step (summarised in \citet{furuta2021co}). 
Model based algorithms such as MD-GPS \citep{montgomery2016guided} and AlphaZero \citep{silver2017mastering} can also be viewed under the EM framework using LQR~\citep{anderson2007optimal} and MCTS~\citep{auer2002finite, coulom2006efficient} during the E-step respectively. \citet{grill2020monte} show MCTS is a form of regularised policy optimisation equivalent to optimising the ELBO. Such model-based algorithms, that iterate between policy improvement using search, and projecting this improvement to the space of parameteriseable policies, are also often referred to as Expert iteration (ExIt)~\citep{silver2017mastering, anthony2017thinking} or Dual Policy Iteration~\citep{sun2018dual}. Such approaches have also been applied to active inference \cite{friston2010free} where MCTS has been used to estimate expected free energy, with this computation then amortised into a neural policy \cite{fountas2020deep}. Our work can also be framed as ExIt, but differs by using Neural SMC for planning with an explicit KL regularisation on the improvement.

Sequential Monte Carlo (SMC)~\citep{gordon1993novel,kitagawa1996monte,liu1998sequential}, also referred to as particle filters~\citep{pitt2001auxiliary, andrieu2004particle}, is an inference method often employed to sample from intractable distributions. 
\citet{gu2015neural} explore the parameterisation of the proposal in SMC using LSTMs~\citep{hochreiter1997long}, showing that posterior estimates from SMC can be used to train a parameterised proposal in non-control based tasks. \citet{li2017approximate} demonstrate the same concept but using an MCMC sampler \citep{gilks1995markov}. 
SMC can be applied to sample from distributions over trajectories in RL. For example, SMC-Learning~\citep{lazaric2007reinforcement} updates SMC weights using a Boltzmann exploration strategy, but unlike SPO it does not control policy improvement with KL regularisation or leverage neural approximators. \citet{piche2018probabilistic} derive an SMC weight update to directly estimate the posterior over optimal trajectories. However this update requires the optimal value function, which they approximate using a SAC~\citep{haarnoja2018soft} value function. This approach requires high sample counts and does not leverage SMC posterior estimates for policy improvement, instead utilising a Gaussian distribution or SAC to train the proposal.
CriticSMC~\citep{lioutas2022critic} estimates the same posterior but utilises a soft action-value function to score particles~\citep{haarnoja2017reinforcement}, reducing the number of steps performed in the environment, enabling more efficient exploration and improved performance on lower SMC budgets compared with~\citet{piche2018probabilistic}. However their method is slower in wall clock time and uses a static proposal, therefore not performing iterative policy improvement. A useful properly of SMC search methods is that it is inherently parallelisable. Parallelising MCTS with a virtual loss has been explored, however this often leads to performance degradation \cite{liu2018watch}, increasing exploration and leading to out-of-distribution states that are difficult to evaluate \cite{dalal2021improve}.

\newpage
\section{Background}
\label{background-RL}

\textbf{Sequential decision-making} can be formalised under the Markov Decision Process (MDP) framework \citep{puterman2014markov}. An MDP is a tuple $(\mathcal{S}, \mathcal{A}, \mathcal{T}, r, \gamma, \mu)$
where, $\mathcal{S}$ is the set of possible states, 
$\mathcal{A}$ is the set of actions, 
$\mathcal{T}: \mathcal{S} \times \mathcal{A}  \rightarrow \mathcal{P}(\mathcal{S})$ is the state transition probability function, 
$r: \mathcal{S} \times \mathcal{A} \rightarrow \mathbb{R}$ is the reward function, 
$\gamma \in [0, 1]$ is the discount factor, and $\mu$ is the initial state distribution. 
An agent interacts with the MDP using a policy $\pi: \mathcal{S} \to \mathcal{P}(\mathcal{A})$, that associates to every state a distribution over actions. We quantify the quality of a policy by the expected discounted return, that the agent seeks to maximise
$ \pi^* \in \arg\max_{\pi \in \Pi} \mathbb{E}_\pi \left[ \sum_{t=0}^{\infty} \gamma^t r_t \right],$
where $r_t = r(s_t, a_t)$ is the reward received at time $t$ and $\Pi$ is the set of all realisable policies. The value function $V^{\pi}(s_t) = \mathbb{E}_\pi\left[ \sum_{t=t}^{\infty} \gamma^t r_t \mid s_t \right]$ maps a state $s_t$ to the expected discounted sum of future rewards when acting according to $\pi$. 
Similarly, the state-action value function $Q^{\pi}(s_t,a_t) = \mathbb{E}_\pi\left[ \sum_{t=t}^{\infty} \gamma^t r_t \mid s_t, a_t \right]$ maps a state $s_t$ to the expected discounted return, when taking the initial action $a_t$ and following $\pi$ thereafter.

\textbf{Control as inference} formulates the RL objective as an inference problem within a probabilistic graphical model~\citep{toussaint2006probabilistic, kappen2012optimal, levine2018reinforcement}. 
For a horizon $T$, the distribution over trajectories $\tau= (s_0, a_0, s_1, a_1, ..., s_T, a_T)$ is given by $p(\tau) = \mu(s_0) \prod_{t=0}^{T} p(a_t)\mathcal{T}(s_{t+1}|s_t, a_t)$, which is a function of the initial state distribution, the transition dynamics, and an action prior. 
Note that this distribution is insufficient for solving control problems, because it has no notion of rewards. We therefore have to introduce an additional \textit{optimality variable} into this model, which we will denote \( \mathcal{O}_t \), that is defined such that $p(\mathcal{O}_t = 1|\tau) \propto \exp(r_t)$. Therefore, a high reward at time $t$ means a high probability of having taken the optimal action at that point. 
We are then concerned with the target distribution $p(\tau | \mathcal{O}_{1:T})$, which is the distribution of trajectories given optimality at every step. We denote $\mathcal{O}_{1:T}$ as $\mathcal{O}$ going forward. 
The RL objective is then formulated as finding a policy to maximise $\log p_{\pi}(\mathcal{O} = 1) = \log \int \pi(\tau) p(\mathcal{O} = 1 | \tau) d\tau$, which intuitively can be thought of as maximising the distribution of optimality at all timesteps, given actions sampled according to $\pi$. To optimise this challenging objective, we can derive the evidence lower bound (ELBO) using an auxiliary distribution~$q$~\citep{neumann2011variational}:
\begin{equation}
\begin{aligned}
\log p_{\pi}(\mathcal{O} = 1) &= \log \int \pi(\tau) p(\mathcal{O} = 1 | \tau) d\tau = \log \int q(\tau) \frac{\pi(\tau) p(\mathcal{O} = 1 | \tau)}{q(\tau)} d\tau \\
&\geq \int q(\tau) \left[ \log p(\mathcal{O} = 1 | \tau) + \log \frac{\pi(\tau)}{q(\tau)} \right] d\tau = \mathbb{E}_q \left[ \sum_t \frac{r_t}{\alpha} \right] - \text{KL}(q(\tau) \| \pi(\tau)).
\label{eq:variational_objective}
\end{aligned}
\end{equation}
Since $ p(O = 1 | \tau) \propto \exp(\sum_t r_t)$ and $\alpha$ is a normalising constant, ensuring a valid probability distribution.

\textbf{Expectation Maximisation }(EM) \citep{dempster1977maximum} has been widely applied to solve the optimisation problem $\max_{\pi} \log p_{\pi}(\mathcal{O} = 1)$ \citep{dayan1997using,peters2007reinforcement,peters2010relative,abdolmaleki2018maximum,song2019v}. 
After deriving the lower bound $\mathcal{J}(q,\pi)$ on the objective in \cref{eq:variational_objective} using an auxiliary non-parametric distribution $q$, EM then performs a coordinate ascent, iterating between optimizing the bound with respect to $q$ (E-step) and with respect to the parametric policy $\pi$ (M-step). 
This generates a sequence of policy pairs $\{(\pi_0,q_0), (\pi_1,q_1), \ldots, (\pi_n, q_n)\}$ such that in each step $i$ in the EM sequence, $\mathcal{J}(q_{i+1},\pi_{i+1}) \geq \mathcal{J}(q_{i},\pi_{i})$. 
Viewed through a traditional policy improvement lens, the E-step corresponds to a policy evaluation phase where we perform rollouts, generating states with their associated estimates for $q$ and value estimates. 
The M-step corresponds to a policy improvement phase where  $\pi$ and $V$ are updated. This step can also be thought of as amortising the probabilistic inference computation performed in the E-step, into a neural network forward pass operation.~\footnote{Many related RL algorithms can be understood by the different approaches to either the E-step or M-step,see \cref{app:em-summary_of_em_methods} for a summary of EM approaches.} 

\textbf{Sequential Monte Carlo} (SMC) methods \citep{gordon1993novel} are designed to sample from an intractable distribution~$p$ referred to as the \textit{target distribution} by using a tractable proposal distribution~$\beta$ (we use $\beta$ to prevent overloading of $q$). 
\textit{Importance Sampling} does this by sampling from $\beta$ and weighting samples by $p(x) / \beta(x)$. The estimate is performed using a set of $N$ particles $\{x^{(n)}\}_{n=1}^N$, where $x^{(n)}$ represents a sample from the support that the target and proposal distributions are defined over, along with the associated importance weights $\{w^{(n)}\}_{n=1}^N$. Each particle uses a sample from the proposal distribution to improve the estimation of the target, increasing $N$ naturally improves the estimation of $p$ \citep{bain2009fundamentals}. 
Once importance weights are calculated, the target is estimated as $\sum_{n=1}^{N} \bar{w}^{(n)} \delta_{x^{(n)}}(x)$, where $\bar{w}$ is the normalised importance sample weight and $\delta_{x^{(n)}}$ is the dirac measure located at $x^{(n)}$. \textit{Sequential Importance Sampling} generalises this for sequential problems, where $x = (x_1, \ldots, x_T)$. In this case importance weights can be calculated iteratively according to:
\begin{align}
w_t(x_{1:t}) = w_{t-1}(x_{1:t-1})\cdot\frac{p(x_t|x_{1:t-1})}{\beta(x_t|x_{1:t-1})}.
\label{SMC:definition}
\end{align}
Particle filtering methods can be affected by \textit{weight degeneracy} \citep{maskell2001tutorial}. This occurs when a few particles dominate the normalised importance weights, rendering the remaining particles negligible. As the variance of particle weights is guaranteed to increase over sequential updates \citep{doucet2000sequential}, this phenomenon is unavoidable. When the majority of particles contribute little to the estimation of the target distribution, computational resources are wasted. \textit{Sequential Importance Resampling} (SIR) \citep{liu2001theoretical} mitigates this problem by periodically resampling particles according to their current weights, subsequently resetting these weights.

\section{SPO Method}

In this section, we present a novel method that combines Sequential Monte Carlo (SMC) sampling with the Expectation Maximisation (EM) framework for policy iteration. We begin by formulating the objective function that we aim to optimise. We then outline our iterative approach to maximising this function, alternating between an expectation step (E-step) and a maximisation step (M-step). Within the E-step, we derive the analytical solution for optimising the objective with respect to the auxiliary distribution $q$. We then demonstrate how SMC can be employed to effectively estimate this \textit{target distribution}. The M-step can be viewed as a projection of the non-parametric policy obtained in the E-step back onto the space of feasible policies. A comprehensive algorithmic outline of our proposed approach is provided in \cref{app:SPO-full-algorithm}.

\subsection{Objective}

We add an additional assumption to the lower bound objective defined in \cref{eq:variational_objective} and assume that the auxiliary distribution $q$, defined over trajectories $\tau$, can be decomposed into individual state dependent distributions, i.e. $q(\tau) = \mu(s_0) \prod_{t \geq 0} q(a_t | s_t) \mathcal{T}(s_{t+1} | s_t, a_t) $. We parameterise $\pi$ using $\theta$ which decomposes in the same way. This enables the objective to be written with respect to individual states instead of full trajectories. Multiplying by $\alpha$, the objective can then be written as follows:
\begin{equation}
\mathcal{J}(q, \pi_\theta) = \mathbb{E}_q \left[ \sum_{t=0}^{\infty} \gamma^t \left[ r_t - \alpha \text{KL}(q(a | s_t) \parallel \pi(a | s_t, \theta)) \right] \right] + \log p(\theta).
\label{eq:optimisation}
\end{equation}

Previous works have explored this formulation where $p(\theta)$ is a prior over the parameters of $\pi$ \cite{hachiya2009efficient,schulman2017equivalence,abdolmaleki2018maximum}.


\subsection{E-step}

Within the expectation step of EM, we maximise \cref{eq:optimisation} with respect to $q$. As in previous work, instead of optimising for the rewards, we consider an objective written according to Q-values \citep{abdolmaleki2018maximum,nair2020awac,liu2022constrained}.
\begin{equation}
\max_q \mathbb{E}_{\mu(s)} \left[ \mathbb{E}_{q(a | s)} \left[ Q^q(s, a) \right] - \alpha \text{KL}(q(\cdot | s) \parallel \pi(\cdot | s, \theta_{i})) \right]
\label{eq:eq4}
\end{equation}
This aims to maximise the expected Q-value of a policy $q$, with the constraint that it doesn't move too far away from $\pi_i$. Framing optimisation with respect to Q-values is useful as it enables the use of function approximators to estimate $Q$.

Maximising this objective is difficult due to the dependence of both the expectation terms and Q-values on $q$. Following previous works, we fix the Q-values with respect to a fixed policy $\bar{\pi}$, resulting in a partial E-step optimisation \citep{silver2017mastering,abdolmaleki2018maximum,song2019v}. We use the most recent estimate of $q$ as the fixed policy we perform maximisation with respect to, similar to AlphaZero's approach of acting according to the expert policy. The initial state distribution $\mu(s)$ is likewise fixed to be distributed according to $\mu_{\bar{\pi}}$. In practice, we use a FIFO replay buffer and sample from it to determine $\mu_{\bar{\pi}}$. 

\Cref{eq:eq4} consists of balancing two objectives. To optimize this, we frame this as a constrained maximization problem, to avoid scaling problems, see \cref{app:constr_scaling} for further information. This objective limits the KL between $q$ and $\pi$ from exceeding a certain threshold:
\begin{equation}
\begin{aligned}
\max_{q} & \ \mathbb{E}_{\mu_{\bar{\pi}}(s)} \left[ \mathbb{E}_{q(a|s)} \left[ Q^{\bar{\pi}}(s, a) \right] \right] \\
\text{s.t.} & \ \mathbb{E}_{\mu_{\bar{\pi}}(s)} \left[ \text{KL} \left( q(a|s) \| \pi(a|s, \theta_i) \right) \right] < \epsilon.
\end{aligned}
\label{eq:constrained_optimisation}
\end{equation}

The following analytic solution to the constrained optimisation problem can then be derived using the Lagrangian multipliers method outlined in \cref{app:em-analytic-derivation}. Likewise $\eta^*$ is obtained by minimising the convex dual function \cref{eq:convex-dual}, and intuitively can be thought of as enforcing the KL constraint on $q$, preventing $q_{i}$ from moving to far from $\pi_i$ :
\begin{equation}
q_{i}(a|s) \propto \pi(a|s, \theta_i) \exp\left(\frac{Q^{\bar{\pi}}(s,a) - V^{\bar{\pi}}(s)}{\eta^*}\right),
\label{eq:a_target}
\end{equation}
where $V^{\bar{\pi}}(s)$ is an action independent baseline. $Q(s,a)-V(s)$ is referred to as the advantage, where optimising for Q-values vs advantages is equivalent \citep{nair2020awac}. Optimising with advantages however, has demonstrated to practically outperform optimising Q-values directly \citep{peng2019advantage,song2019v,nair2020awac}. This is also a natural update to perform as the \textit{policy improvement theorem} \citep{sutton2018reinforcement} outlines that an update by policy iteration can be performed if at least one state-action pair has positive advantage and a non-zero probability of reaching such a state. Although we have a closed form solution for $q$, we do not have either the value function or action-value function in practice. In the next section we outline our approach to estimating this distribution.

\textbf{Regularised Policy Optimisation:} In our closed form solution to the optimisation \cref{eq:a_target}, we are required to solve for $\eta^*$ which ensures that the KL is constrained. 
This can be calculated by minimising the following objective \citep{wirth2016model}, see \cref{app:em-kl-constraining-temp} for derivation: 

\begin{equation}
    g(\eta) = \eta \varepsilon + \eta \int \mu(s) \log \left( \int \pi(a|s, \theta_i) \exp\left(\frac{A^{\bar{\pi}}(s,a)}{\eta}\right) da \right) ds.
\label{eq:convex-dual}
\end{equation}

Practically, we estimate \cref{eq:convex-dual} using a sample-based estimator according to the distribution of states from the replay buffer and stored values for $\pi(a|s, \theta_i)$ and $A^{\bar{\pi}}(s,a)$.


\begin{figure*}[b]
    \centering
    \includegraphics[width=\textwidth]{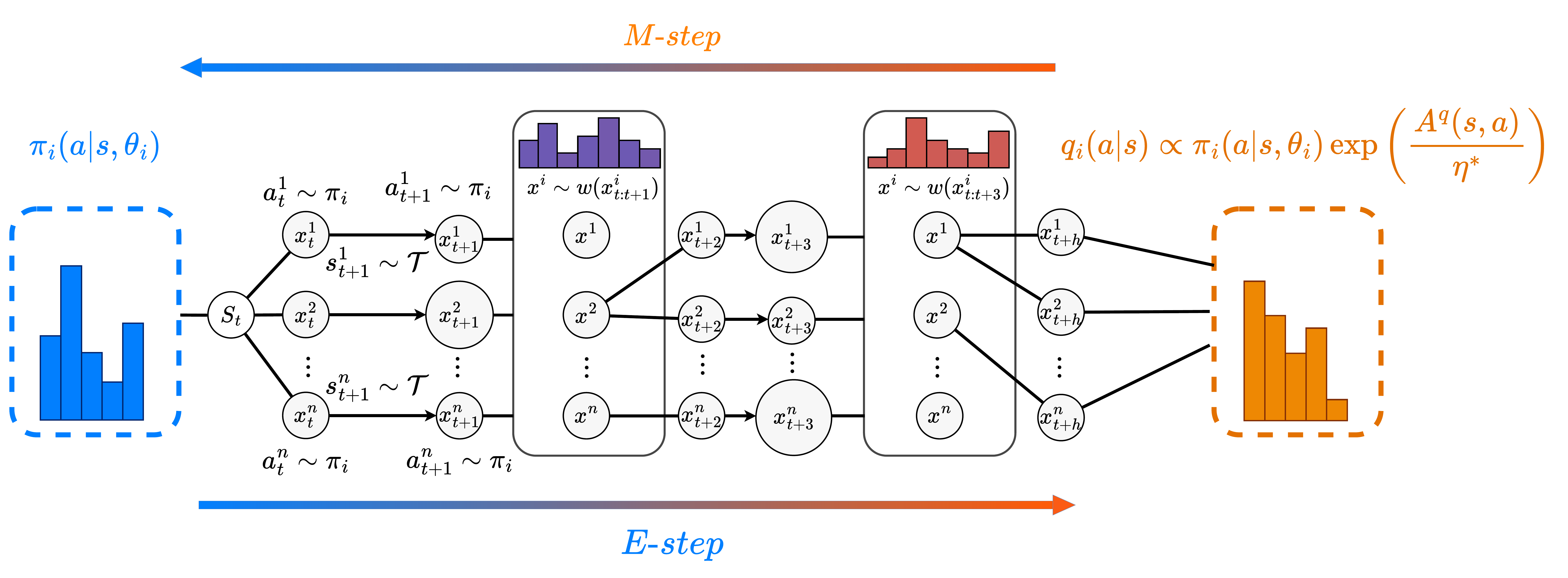}
    \caption{
   SPO search: $n$ rollouts, represented by particles $x^{i}, \dots, x^{n}$, each of which represents an SMC trajectory sample, are performed in parallel according to $\pi_i$ (left to right). At each environment step, the weights of the particles are adjusted (indicated in the diagram by circle size). We show two resampling regions where particles are resampled, favouring those with higher weights, and their weights are reset. The target distribution is estimated from the initial actions of the surviving particles (rightmost particles). This target estimate, $q_{i}$, is then used to update $\pi$ in the M-step.}
    \label{fig:quarks-diagram}
\end{figure*}

\subsubsection{Estimating Target Distribution}

Building upon previous work using Sequential Monte Carlo for trajectory distribution estimation \citep{lazaric2007reinforcement,piche2018probabilistic}, we propose a novel approach that integrates SMC within the Expectation Maximisation framework to estimate the target distribution defined in Equation \ref{eq:a_target}.
Our SMC-based method enables sampling from $q_{i}$ over trajectories of length $h$, incorporating multiple predicted future states and rewards for a more accurate distribution estimation. 
A key property of estimating the target using SMC is that it uses both breadth (we initialise multiple particles to calculate importance weights) but also depth (leveraging future states and rewards) to estimate the distribution. In contrast, MPO evaluates several actions per state, focusing on breadth without depth, while V-MPO calculates advantages using n-step returns from trajectory sequences, leveraging depth but only for a single action sample. Our method combines both aspects, enhancing the accuracy of target distribution estimation, crucial for both action selection and effective policy improvement updates in the M-step. An outline of the SMC algorithm is provided in \cref{alg:smc}, with a visual representation in \cref{fig:quarks-diagram}.

SMC estimates the target by maintaining a set of particles $\{x^{(n)}\}_{n=1}^N$ each of which maintains an importance weight for a particular sample. Given calculated importance weights SMC, estimates the target distribution according to $\hat{q}_{i}{(\tau)} = \sum_{n=1}^{N} \bar{w}^{(n)} \delta_{x_{(n)}}(\tau)$ where $\bar{w}^{(n)}$ is the normalised importance sample weight for a particular sample. We next outline the sequential importance sampling weight update needed to estimate \cref{eq:a_target}. Since we can sample from \(\pi(a_t, s_t, \theta_i)\), we leverage this as our proposal distribution $\beta$. Given the target distribution $q$ can be decomposed into individual state dependent distributions, we can define $p_i(\tau_t|\tau_{1:t-1})$ and $\beta(\tau_{t}|\tau_{1:t-1})$ as:

\begin{equation}
    p_i(\tau_t|\tau_{1:t-1}) \propto \mathcal{T}(s_{t+1}|s_{t},a_{t}) \pi_i(a_t|s_t, \theta_i) \exp\left(\frac{\bar{A_i}(a_t, s_t)}{\eta_{i}^*}\right)
\end{equation}
\begin{equation}
    \beta(\tau_{t}|\tau_{1:t-1}) \propto \mathcal{T}_{\text{model}}(s_{t+1}|s_t,a_t) \pi_i(a_t|s_t, \theta_i)
\end{equation}

leading to the following convenient SMC weight update according to \cref{SMC:definition}:

\begin{equation}
w(\tau_{1:t}) \propto w(\tau_{1:t-1}) \cdot \left(\frac{\mathcal{T}(s_t|s_{t-1},a_{t-1})}{\mathcal{T}_{model}(s_t|s_{t-1},a_{t-1})}\right) \cdot \frac{\exp\left(A^{\bar{\pi}}(a_t, s_t) / \eta^*\right) \cdot \pi(a_t|s_t, \theta_i)}{\pi(a_t|s_t, \theta_i)},
\label{smc:weight_update}
\end{equation}

where \(Q^{\bar{\pi}}(a_t, s_t) - V^{\bar{\pi}}(s_t)\) is simplified to \(A^{\bar{\pi}}(a_t, s_t)\), $\tau_{1:t}$ is a sequence of state, action pairs from timestep $1$ to $t$, and $\mathcal{T}_{model}$ is the environment transition function of the planning model. Note that our work assumes the availability of a model that accurately represents the transition dynamics of the environment \(\mathcal{T}\), and therefore simplify the update to $w(\tau_{1:t}) \propto w(\tau_{1:t-1}) \cdot \exp\left(A^{\bar{\pi}}(a_t, s_t) / \eta^*\right) $.

\Cref{alg:smc} outlines the method for estimating \cref{eq:a_target} over $h$ planning steps for $N$ particles (line 3) using the advantage based weight update (line 7) in \cref{smc:weight_update}. Once $h$ steps in the environment model have been performed in parallel, we marginalise all but the first actions as a sample based estimate of $q_{i}$ (line 14). 

\begin{wrapfigure}[17]{r}{0.5\textwidth} 
\setlength{\intextsep}{-15pt} 
\setlength{\columnsep}{0pt} 
\begin{minipage}{0.5\textwidth}
\begin{algorithm}[H]
\small
\caption{SMC $q$ target estimation (timestep $t$)}
\begin{algorithmic}[1]
\label{alg:smc}
\State Initialize $\{s_t^{(n)} = s_t\}_{n=1}^N$
\State Set $\{w_t^{(n)} = 1\}_{n=1}^N$
\For{$i \in \{t+1, \ldots, t + h\}$}
    \State $\{a_i^{(n)} \sim \pi(\cdot|s_i^{(n)},\theta)\}_{n=1}^N$
    \State $\{s_{i+1}^{(n)} \sim \mathcal{T}_{model}(s_i^{(n)}, a_i^{(n)})\}_{n=1}^N$
    \State $\{r_i^{(n)} \sim r_{model}(s_i^{(n)}, a_i^{(n)})\}_{n=1}^N$
    \State $\{w_i^{(n)} = w_{i-1}^{(n)} \cdot \exp(\hat{A}(s_i^{n},r_i^{n},s_{i+1}^{n})/ \eta^*)\}_{n=1}^N$
    \If{$(i - t) \bmod p = 0$} 
        \State $\{x_{t:i}^{(n)}\}_{n=1}^N \sim \text{Mult}(n; w_i^{(1)},.., w_i^{(N)})$
        \State $\{w_{i}^{(n)} = 1\}_{n=1}^N$
    \EndIf
\EndFor
\State $\{a_t^{(n)}\}$ as the set of first actions of $\{x_{t:t+h}^{(n)}\}_{n=1}^N$
\State $\hat{q}(a|s_t) = \sum_{n=1}^{N} \bar{w}^{(n)}\delta_{a_t^{(n)}}(a)$ 
\end{algorithmic}
\end{algorithm}
\end{minipage}
\end{wrapfigure}

The advantage function $A^{\bar{\pi}}$ is typically unknown, so we compute a 1-step estimate at each iteration using the value function and observed environment rewards $\hat{A}(s_t,a) = r_t + V^{\bar{\pi}}(s_{t+1}) - V^{\bar{\pi}}(s_t)$. Practically, we parameterise the value function $V$ using a neural network and train it using GAE \citep{schulman2016high} on true environment rollouts collected. After $h$ steps, the importance weights leverage $h$-steps of observed states and rewards during planning and corresponding advantage estimates. Compared to tree-based methods such as MCTS, SMC does not require maintaining the full tree in memory. Instead, after each planning step, it only needs to retain the initial action, current state, and current particle weight. It also does not require spending computation sending backward messages to update statistics of previous nodes everytime a new node is added to the tree.

\textbf{Resampling Adaptive Search:} To mitigate the issue of \textit{weight degeneracy} \citep{maskell2001tutorial}, SPO conducts periodic resampling (lines 8-11). This involves generating a new set of particles from the existing set by duplicating some trajectories and removing others, based on the current particle weights. This process enhances computational efficiency in estimating the target distribution. By resampling, we avoid updating weights for trajectories with low likelihood under the target, thereby reallocating computational resources to particles with high importance weights \citep{doucet2001sequential}.


\subsection{M-step}
After completing the E-step (which generates $q_{i}$), we proceed with the M-step, which optimises \cref{eq:optimisation} with respect to $\pi$, parametrised by $\theta$. By eliminating terms that are independent of $\pi$, we optimise the following objective, corresponding to a maximum a posteriori estimation with respect to the distribution $q_{i}$:
\begin{equation}
\max_{\theta} \mathcal{J}(q_i,\pi_\theta) = \max_{\theta} \mathbb{E}_{\mu_{q_i}(s)} \left[ \mathbb{E}_{q_{i}( \cdot |s)} \left[ \log \pi(a|s, \theta) \right] \right] + \log p(\theta).
\label{eq:m-step-pre-step}
\end{equation}

This optimisation can be viewed as projecting the non-parametric policy $q_i$ back to the space of parametrisable policies $\Pi_\theta$, as performed in expert iteration style methods such as AlphaZero. $p(\theta)$ represents a prior over the parameter $\theta$. Previous works find that utilising a prior for $\theta$ to be close to the estimate from the previous iteration $\theta_{i}$ leads to stable training \citep{abdolmaleki2018maximum,song2019v}. Therefore we assume a gaussian prior over the current policy parameters, see \cref{app:fisher} where we show utilising such a prior leads to the following constrained objective: 
\begin{equation}
\begin{aligned}
\max_{\theta}  & \  \mathbb{E}_{\mu_{q_{i}}(s)} \left[ \mathbb{E}_{q_{i}(a|s)} \left[ \log \pi(a|s, \theta) \right] \right] \\
\text{s.t.} & \quad \mathbb{E}_{\mu_{q_{i}}(s)} \left[ \text{KL} \left( \pi(a|s, \theta_{i}), \pi(a|s, \theta) \right) \right] < \epsilon_m.
\end{aligned}
\label{eq:m-step}
\end{equation}

\subsection{Policy Improvement}

Previous approaches to using SMC for RL either do not perform iterative policy improvement using SMC \citep{piche2018probabilistic}, or lack policy improvement constraints needed for iterative improvement \citep{lazaric2007reinforcement}. Assuming that SMC provides perfect estimates of the analytic \textit{target distribution} at each step $i$, Expectation Maximisation algorithm will guarantee that successive iterations will result in monotonic improvements in our lower bound objective, with details outlined in \cref{app:em-monotone-improvement} \citep{liu2022constrained}. 

\begin{proposition}
Given a non-parametric variational distribution $q_i$ and a parametric policy $\pi_{\theta_i}$. Given $q_{i+1}$, the analytical solution to E-step optimisation \cref{eq:optimisation} , and $\pi_{\theta_{i+1}}$, the solution to maximisation problem in the M-step \cref{eq:m-step} then the ELBO $\mathcal{J}$ is guaranteed to be monotonically increasing: $\mathcal{J}(q_{i+1}, \pi_{\theta_{i+1}}) \geq \mathcal{J}(q_{i}, \pi_{\theta_i})$.
\label{eq:em-monotone-improvement}
\end{proposition}

In practice we are unlikely to generate perfect estimates of the target through sample based inference and leave derivations regarding the impact of this estimation on overall convergence for future work. We also draw the connection between our EM optimisation method and Mirror Descent Guided Policy Search (MD-GPS) \citep{montgomery2016guided}. Our objective can be viewed as a specific instance of MD-GPS (see \cref{app:md-gps}). Depending on whether dynamics are linear or not, optimising the EM objective can be viewed either as exact or approximate mirror descent~\citep{beck2003mirror}. Monotonic improvement guarantees in MD-GPS follow from those of mirror descent.
\section{Experiments}

In this section, we focus on three main areas of analysis. First, we demonstrate the improved performance of SPO in terms of episode returns, relative to both model-free and model-based algorithms. We conduct evaluations across a suite of common environments for both continuous control and discrete action spaces. Secondly, we examine the scaling behaviour of SPO during training, showing that asymptotic performance scales with particle count and depth. Finally, we explore the performance-to-speed trade-off at test time by comparing SPO directly to AlphaZero as the search budget increases.

\subsection{Experimental Setup} 
In order to ensure the robustness of our conclusions, we follow the evaluation methodology proposed by \citet{agarwal2021deep}, see \cref{app:stat-prec} for further details. This evaluation methodology groups performance across tasks within an environment suite, enabling clearer conclusions over the significance of learning algorithms as a whole. We include individual results in \cref{app:expanded-results}, along with additional analysis measuring the statistical significance of our results.

\textbf{Environments:} 
For continuous control we evaluate on the Brax \citep{brax2021github} benchmark environments of: Ant, HalfCheetah, and Humanoid. 
For discrete environments, we evaluate on Boxoban \citep{boxobanlevels} (a specific instance of Sokoban), commonly used to assess planning methods, and Rubik's Cube, a sparse reward environment with a large combinatorial state-action space. See \cref{app:environments} for further details regarding environments.

\textbf{Baselines:} 
Our model-free baselines include PPO \citep{schulman2017proximal}, MPO \citep{abdolmaleki2018maximum} and V-MPO \citep{song2019v} for both continuous and discrete environments. 
For model-based algorithms, we compare performance to AlphaZero (including search improvements from MuZero~\citep{schrittwieser2020mastering}) and an SMC method introduced by \citet{piche2018probabilistic}~\footnote{We use MDQN \citep{vieillard2020munchausen}, and SAC \citep{haarnoja2018soft} for discrete and continuous environments, respectively. Note that in the discrete case, SMC-ENT using SAC was very unstable, so we chose MDQN as a better alternative.} (which we refer to as SMC-ENT due to its use of maximum entropy RL to train the proposal and value function used within SMC). Our AlphaZero benchmark follows the official open-source implementation\footnote{Open-source implementation available at \href{https://github.com/google-deepmind/mctx/}{mctx}.}. For continuous environments we baseline our results to Sampled MuZero \citep{hubert2021learning}, a modern extension to MuZero for large and/or continuous action spaces. We utilise a true environment model, aligning our implementation of Sampled MuZero more closely with AlphaZero. Our core experiments configure SPO with 16 particles and a horizon of 4, and AlphaZero with 64 simulations to equalise search budgets. For remaining parameters, see \cref{app:baseline-hyperparameters} and \cref{app:spo-hyperparameters}. Each environment and algorithm were evaluated with five random seeds.

\begin{figure*}[t!]
    \centering
    \begin{minipage}{0.49\textwidth}
        \centering
        \includegraphics[width=\linewidth]{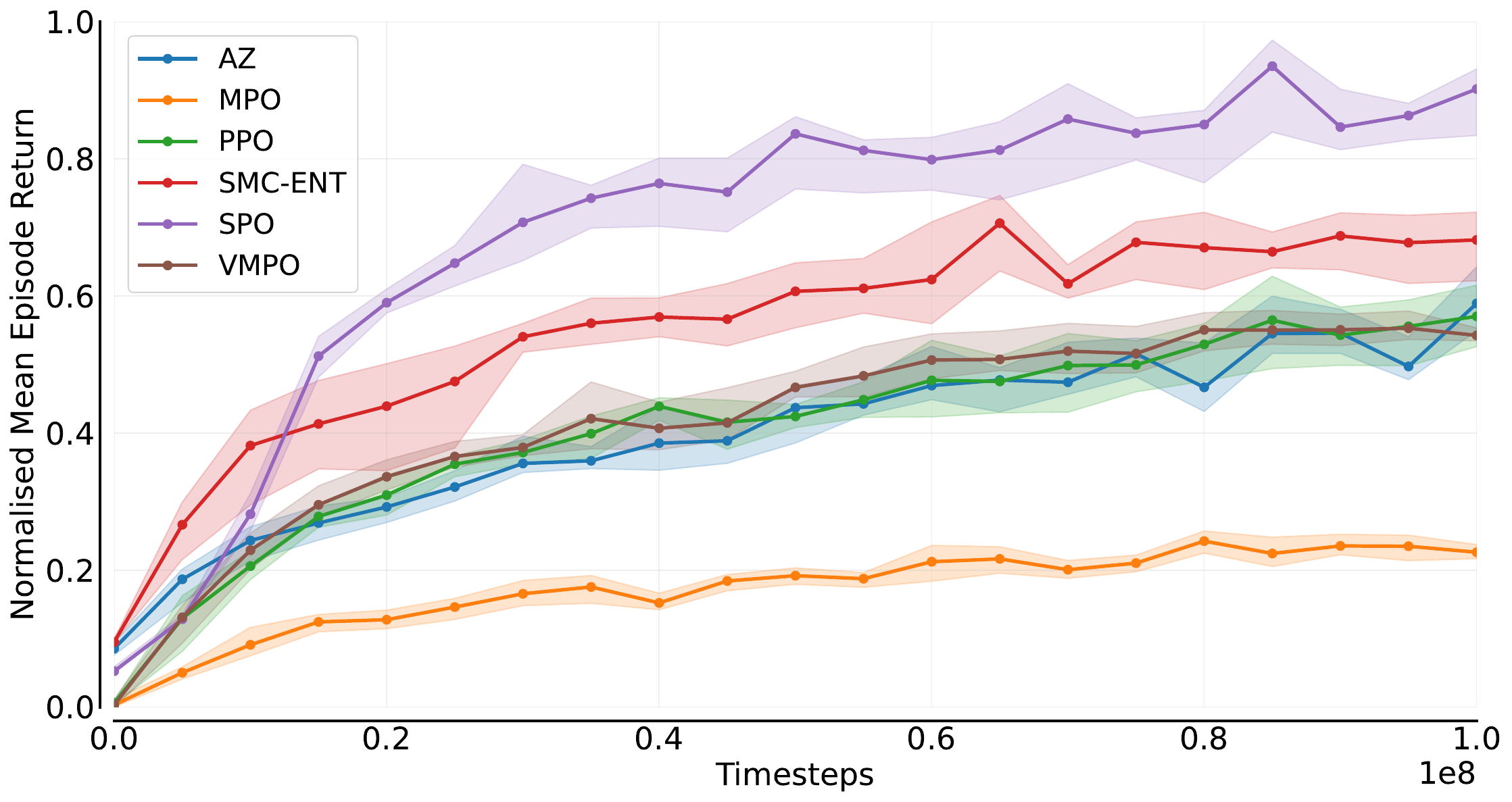}
        \textbf{(a) Discrete:} Rubik's Cube 7 and Boxoban Hard.
    \end{minipage}%
    \hfill
    \begin{minipage}{0.49\textwidth}
        \centering
        \includegraphics[width=\linewidth]{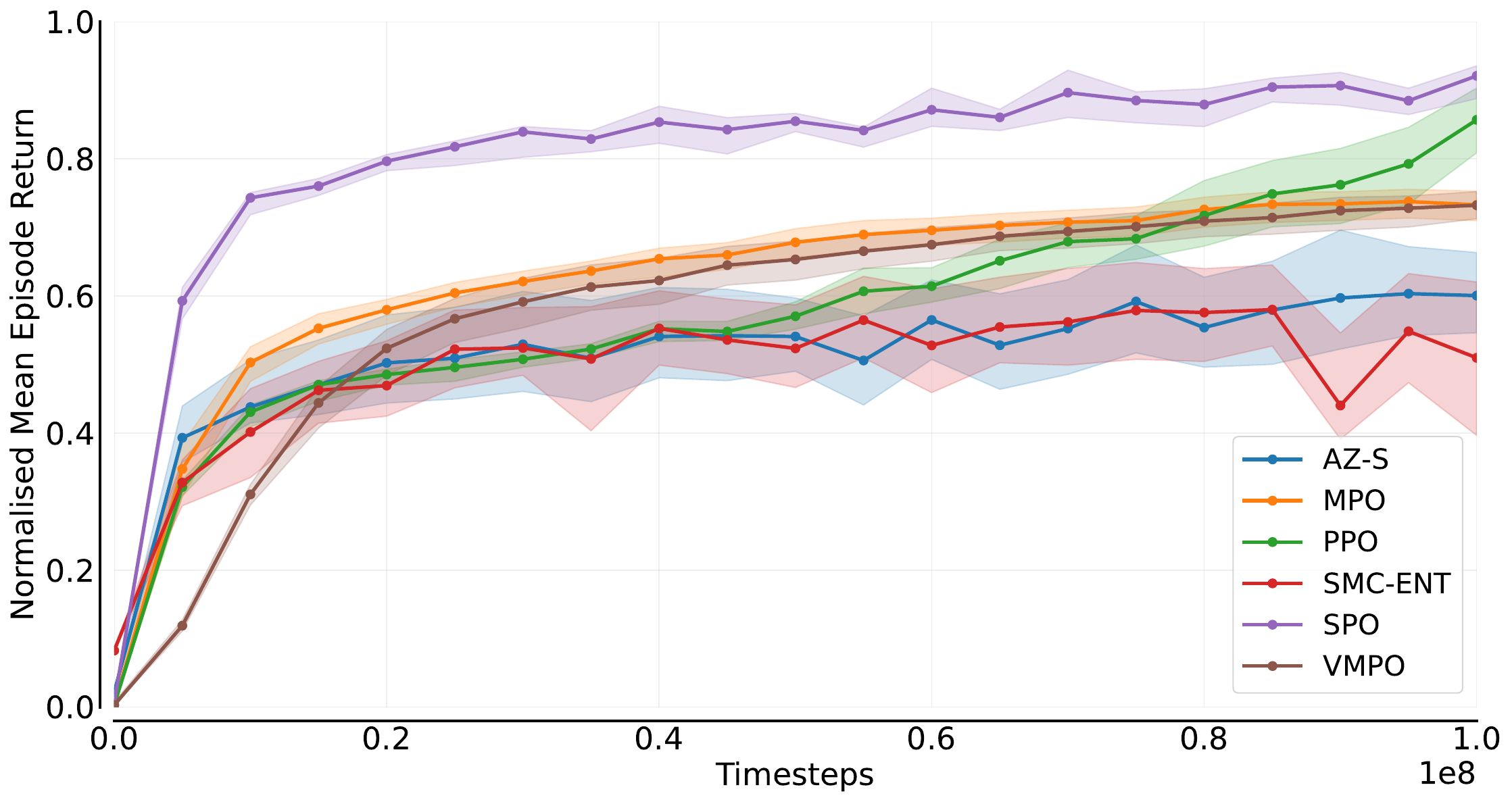}
        \textbf{(b) Continuous:} Ant, HalfCheetah, Humanoid.
    \end{minipage}
    \caption{Learning curves for discrete and continuous environments. The Y-axis represents the interquartile mean of min-max normalised scores, with shaded regions indicating 95\% confidence intervals, across 5 random seeds.}
    \label{fig:training-curve-results}
\end{figure*}

\subsection{Results}\label{sec:results:perf}

\textbf{Policy Improvement:} In \cref{fig:training-curve-results} we show empirical evidence that SPO outperforms all baseline methods across both discrete (left) and continuous (right) benchmarks for policy improvement. This conclusion also holds for the per-environment results reported in \cref{app:expanded-results}. 
SMC-ENT \citep{piche2018probabilistic} is a strong SMC based algorithm, however, SPO outperforms it for both discrete and continuous environments providing strong evidence of the direct benefit of using SMC posterior estimates for policy improvement. We also highlight the importance regularising policy optimisation, which is practically achieved by solving for $\eta^*$  \cref{eq:convex-dual}, and is re-estimated at each iteration. In \cref{app:ablations} we ablate the impact of this adaptive method by comparing varying fixed temperature schedules. While a good temperature can be found through expensive grid search for each new environment we find the adaptive temperature ensures stable updates leading to strong learning performance across all environments, with minimal overhead to compute $\eta^*$.

We find that SPO performance is statistically significant compared to AlphaZero across the evaluated environments, see \cref{app:expanded-results} for detailed results. The substantial variation in AlphaZero performance across these environments underscores the difficulty in tuning it for diverse settings. For instance, while AlphaZero performs well on Boxoban and HalfCheetah, its performance drops significantly on other discrete and continuous problems. This inconsistency poses a major challenge for its applicability to real-world problems. In contrast, SPO performs consistently well across all environments (both discrete and continuous), highlighting its general applicability and robustness to various problem settings.

\subsection{Scaling SPO during training}\label{sec:results:scaling}

\begin{figure*}[t!]
  \centering
  \includegraphics[width=0.49\linewidth]{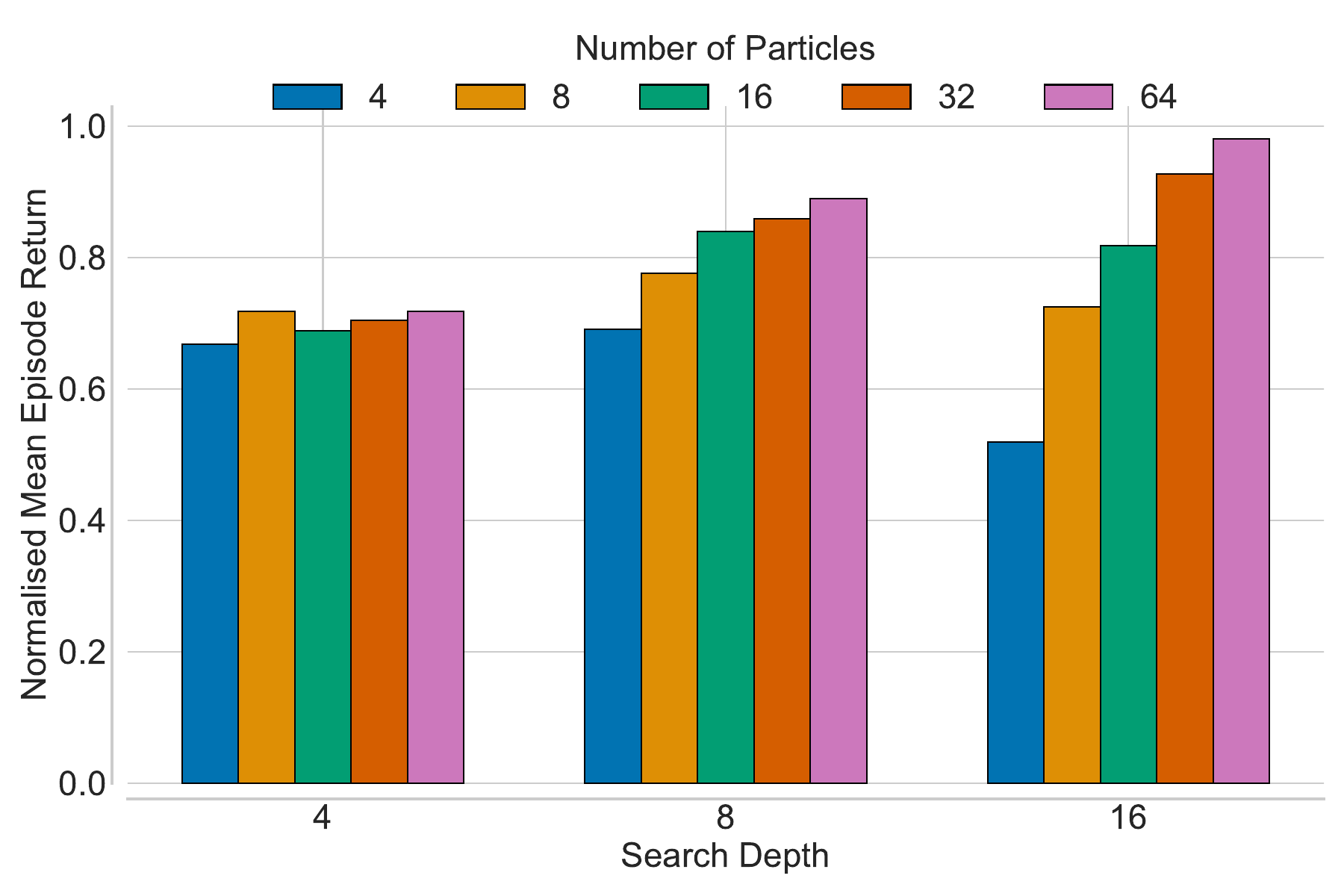}
  \includegraphics[width=0.49\linewidth]{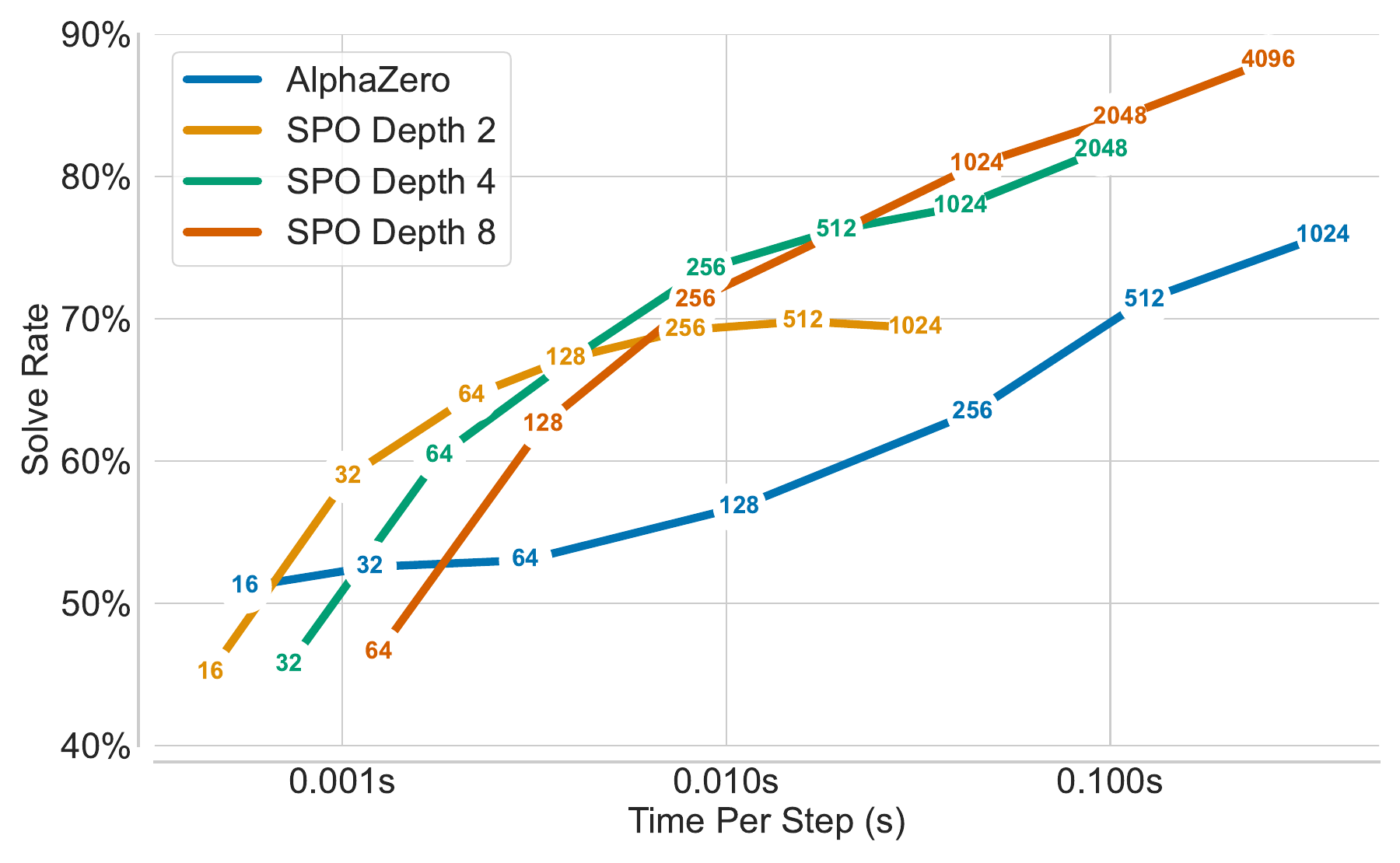}
  \caption{
  (left) Scaling: Mean normalised performance across all continuous environments on $10^8$ environment steps, varying particle numbers $N$ and horizon $h$ for SPO during training.
  (right) Wall Clock Time Comparison: Performance on Rubik's cube plotted against wall-clock time for AlphaZero and 3 versions of SPO (varying by SMC search depth), with total search budget labeled at each point.}
  \label{fig:scaling-perf}
\end{figure*}

In \cref{fig:scaling-perf} (left) we investigate how SPO performance is impacted by scaling particle counts $N$ and horizon length $h$ during training, both of which we find improve the estimation of the \textit{target distribution}. Our results show that scaling both the particle count and the horizon leads to improvements in the asymptotic performance of SPO. It also suggests that both variables should be scaled together for maximum benefit as having a long horizon with low particle counts can actually negatively impact performance. 
Secondly, we highlight that while for our main results we use a total search budget (particles $\times$ depth) of $64$ for policy improvement, our scaling results show that competitive results can be achieved by reducing this budget by a factor of four, which demonstrates competitive performance at low compute budgets.

\subsection{Scaling SPO at test time}\label{sec:results:speed}
We can scale particle counts and horizon during training to improve target distribution estimates for the M-step. However, this scaling can also be performed after training when $\pi_{\theta_i}$ is fixed, enhancing performance, since actions are sampled in the environment according to the improved policy $q$. This equates to additional E-step optimization, which can enhance performance due to the representational constraints of the space of parameterised policies. In \cref{fig:scaling-perf} (right) we demonstrate how the performance of both SPO and AlphaZero search scales with additional search budget at test time using the same high performing checkpoint. We plot the time taken for a single actor step (measured on a TPUv3-8) against solve rate for the Rubik's Cube problem with time on a logarithmic axis. Specifically, we evaluate on 1280 episodes for cubes ten scrambles away from solved. While we recognise that such analysis can be difficult to perform due to implementation discrepancies, we used the official JAX implementation of AlphaZero (MCTX) within our codebase and compare this to SPO. Additionally, we exclusively measure inference thus no training implementation details affect our measurements.

This provides evidence that AlphaZero has worse scaling when compared to SPO on horizons four and eight, noting that for horizon of two, SPO performance converges early, as low depths can act as a bottleneck for further performance improvements. This highlights the benefits of the SPO parallelism. This chart also shows how for different compute preferences at test time, a different horizon length is preferable. For limited compute, low horizon lengths and relatively higher particle counts provide the best performance, but as compute availability increases, the gains from increasing particle counts decrease and instead horizon length $h$ should be increased.

Lastly, the plot illustrates the increase in available search budget, by using SPO, given a time restriction rather than a compute restriction. For example, SPO can use 4 times more search budget, when compared to AlphaZero, given a requirement of 0.1 second per step.

\section{Conclusion} Planning-based policy improvement operators have proven to be powerful methods to enhance the learning of policies for complex environments when compared to model-free methods. Despite the success of these methods, they are still limited in their usefulness, due to the requirement of high planning budgets and the need for algorithmic modifications to adapt to large or continuous action spaces. In this paper, we introduce a modern implementation of Sequential Monte Carlo planning as a policy improvement operator, within the Expectation Maximisation framework. This results in a general training methodology with the ability to scale in both discrete and continuous environments.

Our work provides several key contributions. First, we show that SPO is a powerful policy improvement operator, outperforming our model-based and model-free baselines. Additionally, SPO is shown to be competitive across both continuous and discrete domains, without additional environment-specific alterations. This illustrates the effectiveness of SPO as a generic and robust approach, in contrast to prior methods that require domain-specific enhancements. Furthermore, the parallelisable nature of SPO results in efficient scaling behaviour of search. This allows SPO to achieve a significant boost in performance at inference time by scaling the search budget. Finally, we demonstrate that scaling SPO results in faster wall-clock inference time compared to previous work utilising tree-based search methods. The presented work culminates in a versatile, neural-guided, sample-based planning method that demonstrates superior performance and scalability over baselines.

\textbf{Limitations \& Future Work:} This work demonstrates the efficacy of combining probabilistic inference with reinforcement learning and amortisation. While our work considers a relatively standard form of Sequential Monte Carlo, future work could investigate making improvements to this inference method, in order to further improve the estimate of the target distribution, while maintaining the scalability benefits demonstrated. We also draw the connection to Active Inference \cite{friston2010free} which has been explored in the context of agent behaviour in complex environments, where Monte Carlo Tree Search has been used to scale previous methods along with deep learning \cite{fountas2020deep}. Leveraging Sequential Monte Carlo in such methods along with amortisation could be a promising direction of research to further scale such methods. Our work exclusively uses exact world models of the environment in order to isolate the effects of various planning methods on performance. Extending our research to include a learned world model would broaden the applicability of SPO to more complex problems that may not have a perfect simulator and are therefore unsuitable for direct planning. Additionally, we apply Sequential Monte Carlo (SMC) only in deterministic settings. Adapting our approach for stochastic environments presents challenges. Specifically, the importance weight update in SMC can lead to the selection of dynamics most advantageous to the agent, potentially fostering risk-seeking behaviour in stochastic settings. However, mitigation strategies exist, as discussed in \citet{levine2018reinforcement}. 
\begin{ack}
The authors would like to thank Teodora Pandeva for detailed discussions on SPO theory and for detailed revisions of the manuscript. Cl\'ement Bonnet for brainstorming discussions and feedback and David Kuric for revisions of the final manuscript. We also thank members of the InstaDeep team for their support in the preparation of the final manuscript. Lastly we thank the anonymous reviewers for comments and helpful discussions that helped improve the final version of the paper.

Research was supported with Cloud TPUs from Google’s TPU Research Cloud (TRC). Matthew Macfarlane is supported by the LIFT-project 019.011, which is partly financed by the Dutch Research Council (NWO).

\end{ack}


\bibliographystyle{plainnat} 
\bibliography{main}

\begin{thebibliography}{84}
\providecommand{\natexlab}[1]{#1}
\providecommand{\url}[1]{\texttt{#1}}
\expandafter\ifx\csname urlstyle\endcsname\relax
  \providecommand{\doi}[1]{doi: #1}\else
  \providecommand{\doi}{doi: \begingroup \urlstyle{rm}\Url}\fi

\bibitem[Abdolmaleki et~al.(2018)Abdolmaleki, Springenberg, Tassa, Munos, Heess, and Riedmiller]{abdolmaleki2018maximum}
Abbas Abdolmaleki, Jost~Tobias Springenberg, Yuval Tassa, Remi Munos, Nicolas Heess, and Martin Riedmiller.
\newblock Maximum a posteriori policy optimisation.
\newblock In \emph{International Conference on Learning Representations}, 2018.
\newblock URL \url{https://openreview.net/forum?id=S1ANxQW0b}.

\bibitem[Adkins et~al.()Adkins, Bowling, and White]{adkinsmethod}
Jacob Adkins, Michael Bowling, and Adam White.
\newblock A method for evaluating hyperparameter sensitivity in reinforcement learning.
\newblock In \emph{Finding the Frame: An RLC Workshop for Examining Conceptual Frameworks}.

\bibitem[Agarwal et~al.(2021)Agarwal, Schwarzer, Castro, Courville, and Bellemare]{agarwal2021deep}
Rishabh Agarwal, Max Schwarzer, Pablo~Samuel Castro, Aaron~C Courville, and Marc Bellemare.
\newblock Deep reinforcement learning at the edge of the statistical precipice.
\newblock \emph{Advances in neural information processing systems}, 34:\penalty0 29304--29320, 2021.

\bibitem[Amos et~al.(2023)]{amos2023tutorial}
Brandon Amos et~al.
\newblock Tutorial on amortized optimization.
\newblock \emph{Foundations and Trends{\textregistered} in Machine Learning}, 16\penalty0 (5):\penalty0 592--732, 2023.

\bibitem[Anderson and Moore(2007)]{anderson2007optimal}
Brian~DO Anderson and John~B Moore.
\newblock \emph{Optimal control: linear quadratic methods}.
\newblock Courier Corporation, 2007.

\bibitem[Andrieu et~al.(2004)Andrieu, Doucet, Singh, and Tadic]{andrieu2004particle}
Christophe Andrieu, Arnaud Doucet, Sumeetpal~S Singh, and Vladislav~B Tadic.
\newblock Particle methods for change detection, system identification, and control.
\newblock \emph{Proceedings of the IEEE}, 92\penalty0 (3):\penalty0 423--438, 2004.

\bibitem[Anthony et~al.(2017)Anthony, Tian, and Barber]{anthony2017thinking}
Thomas Anthony, Zheng Tian, and David Barber.
\newblock Thinking fast and slow with deep learning and tree search.
\newblock \emph{Advances in neural information processing systems}, 30, 2017.

\bibitem[Antonoglou et~al.(2021)Antonoglou, Schrittwieser, Ozair, Hubert, and Silver]{antonoglou2021planning}
Ioannis Antonoglou, Julian Schrittwieser, Sherjil Ozair, Thomas~K Hubert, and David Silver.
\newblock Planning in stochastic environments with a learned model.
\newblock In \emph{International Conference on Learning Representations}, 2021.

\bibitem[Auer et~al.(2002)Auer, Cesa-Bianchi, and Fischer]{auer2002finite}
Peter Auer, Nicolo Cesa-Bianchi, and Paul Fischer.
\newblock Finite-time analysis of the multiarmed bandit problem.
\newblock \emph{Machine learning}, 47\penalty0 (2):\penalty0 235--256, 2002.

\bibitem[Bain and Crisan(2009)]{bain2009fundamentals}
Alan Bain and Dan Crisan.
\newblock \emph{Fundamentals of stochastic filtering}, volume~3.
\newblock Springer, 2009.

\bibitem[Beck and Teboulle(2003)]{beck2003mirror}
Amir Beck and Marc Teboulle.
\newblock Mirror descent and nonlinear projected subgradient methods for convex optimization.
\newblock \emph{Operations Research Letters}, 31\penalty0 (3):\penalty0 167--175, 2003.

\bibitem[Bonnet et~al.(2023)Bonnet, Luo, Byrne, Surana, Duckworth, Coyette, Midgley, Abramowitz, Tegegn, Kalloniatis, et~al.]{bonnet2023jumanji}
Cl{\'e}ment Bonnet, Daniel Luo, Donal~John Byrne, Shikha Surana, Paul Duckworth, Vincent Coyette, Laurence~Illing Midgley, Sasha Abramowitz, Elshadai Tegegn, Tristan Kalloniatis, et~al.
\newblock Jumanji: a diverse suite of scalable reinforcement learning environments in jax.
\newblock In \emph{The Twelfth International Conference on Learning Representations}, 2023.

\bibitem[Bouthillier et~al.(2021)Bouthillier, Delaunay, Bronzi, Trofimov, Nichyporuk, Szeto, Mohammadi~Sepahvand, Raff, Madan, Voleti, et~al.]{bouthillier2021accounting}
Xavier Bouthillier, Pierre Delaunay, Mirko Bronzi, Assya Trofimov, Brennan Nichyporuk, Justin Szeto, Nazanin Mohammadi~Sepahvand, Edward Raff, Kanika Madan, Vikram Voleti, et~al.
\newblock Accounting for variance in machine learning benchmarks.
\newblock \emph{Proceedings of Machine Learning and Systems}, 3:\penalty0 747--769, 2021.

\bibitem[Bradbury et~al.(2018)Bradbury, Frostig, Hawkins, Johnson, Leary, Maclaurin, Necula, Paszke, Vander{P}las, Wanderman-{M}ilne, and Zhang]{jax2018github}
James Bradbury, Roy Frostig, Peter Hawkins, Matthew~James Johnson, Chris Leary, Dougal Maclaurin, George Necula, Adam Paszke, Jake Vander{P}las, Skye Wanderman-{M}ilne, and Qiao Zhang.
\newblock {JAX}: composable transformations of {P}ython+{N}um{P}y programs, 2018.
\newblock URL \url{http://github.com/google/jax}.

\bibitem[Colas et~al.(2018)Colas, Sigaud, and Oudeyer]{colas2018gep}
C{\'e}dric Colas, Olivier Sigaud, and Pierre-Yves Oudeyer.
\newblock Gep-pg: Decoupling exploration and exploitation in deep reinforcement learning algorithms.
\newblock In \emph{International conference on machine learning}, pages 1039--1048. PMLR, 2018.

\bibitem[Coulom(2006)]{coulom2006efficient}
R{\'e}mi Coulom.
\newblock Efficient selectivity and backup operators in {Monte-Carlo} tree search.
\newblock In \emph{International conference on computers and games}, pages 72--83. Springer, 2006.

\bibitem[Dalal et~al.(2021)Dalal, Hallak, Dalton, Mannor, Chechik, et~al.]{dalal2021improve}
Gal Dalal, Assaf Hallak, Steven Dalton, Shie Mannor, Gal Chechik, et~al.
\newblock Improve agents without retraining: Parallel tree search with off-policy correction.
\newblock \emph{Advances in Neural Information Processing Systems}, 34:\penalty0 5518--5530, 2021.

\bibitem[Danihelka et~al.(2021)Danihelka, Guez, Schrittwieser, and Silver]{danihelka2021policy}
Ivo Danihelka, Arthur Guez, Julian Schrittwieser, and David Silver.
\newblock Policy improvement by planning with gumbel.
\newblock In \emph{International Conference on Learning Representations}, 2021.

\bibitem[Dayan and Hinton(1997)]{dayan1997using}
Peter Dayan and Geoffrey~E Hinton.
\newblock Using expectation-maximization for reinforcement learning.
\newblock \emph{Neural Computation}, 9\penalty0 (2):\penalty0 271--278, 1997.

\bibitem[Dempster et~al.(1977)Dempster, Laird, and Rubin]{dempster1977maximum}
Arthur~P Dempster, Nan~M Laird, and Donald~B Rubin.
\newblock Maximum likelihood from incomplete data via the em algorithm.
\newblock \emph{Journal of the royal statistical society: series B (methodological)}, 39\penalty0 (1):\penalty0 1--22, 1977.

\bibitem[Doucet et~al.(2000)Doucet, Godsill, and Andrieu]{doucet2000sequential}
Arnaud Doucet, Simon Godsill, and Christophe Andrieu.
\newblock On sequential monte carlo sampling methods for bayesian filtering.
\newblock \emph{Statistics and computing}, 10:\penalty0 197--208, 2000.

\bibitem[Doucet et~al.(2001)Doucet, De~Freitas, Gordon, et~al.]{doucet2001sequential}
Arnaud Doucet, Nando De~Freitas, Neil~James Gordon, et~al.
\newblock \emph{Sequential Monte Carlo methods in practice}, volume~1.
\newblock Springer, 2001.

\bibitem[Dror et~al.(2019)Dror, Shlomov, and Reichart]{dror2019deep}
Rotem Dror, Segev Shlomov, and Roi Reichart.
\newblock Deep dominance-how to properly compare deep neural models.
\newblock In \emph{Proceedings of the 57th Annual Meeting of the Association for Computational Linguistics}, pages 2773--2785, 2019.

\bibitem[Fawzi et~al.(2022)Fawzi, Balog, Huang, Hubert, Romera-Paredes, Barekatain, Novikov, Ruiz, Schrittwieser, Swirszcz, Silver, Hassabis, and Kohli]{AlphaTensor2022}
Alhussein Fawzi, Matej Balog, Aja Huang, Thomas Hubert, Bernardino Romera-Paredes, Mohammadamin Barekatain, Alexander Novikov, Francisco J.~R. Ruiz, Julian Schrittwieser, Grzegorz Swirszcz, David Silver, Demis Hassabis, and Pushmeet Kohli.
\newblock Discovering faster matrix multiplication algorithms with reinforcement learning.
\newblock \emph{Nature}, 610\penalty0 (7930):\penalty0 47--53, 2022.
\newblock \doi{10.1038/s41586-022-05172-4}.

\bibitem[Fountas et~al.(2020)Fountas, Sajid, Mediano, and Friston]{fountas2020deep}
Zafeirios Fountas, Noor Sajid, Pedro Mediano, and Karl Friston.
\newblock Deep active inference agents using monte-carlo methods.
\newblock \emph{Advances in neural information processing systems}, 33:\penalty0 11662--11675, 2020.

\bibitem[Freeman et~al.(2021)Freeman, Frey, Raichuk, Girgin, Mordatch, and Bachem]{brax2021github}
C.~Daniel Freeman, Erik Frey, Anton Raichuk, Sertan Girgin, Igor Mordatch, and Olivier Bachem.
\newblock Brax - a differentiable physics engine for large scale rigid body simulation, 2021.
\newblock URL \url{http://github.com/google/brax}.

\bibitem[Friston(2010)]{friston2010free}
Karl Friston.
\newblock The free-energy principle: a unified brain theory?
\newblock \emph{Nature reviews neuroscience}, 11\penalty0 (2):\penalty0 127--138, 2010.

\bibitem[Furmston and Barber(2010)]{furmston2010variational}
Thomas Furmston and David Barber.
\newblock Variational methods for reinforcement learning.
\newblock In \emph{Proceedings of the Thirteenth International Conference on Artificial Intelligence and Statistics}, pages 241--248. JMLR Workshop and Conference Proceedings, 2010.

\bibitem[Furuta et~al.(2021)Furuta, Kozuno, Matsushima, Matsuo, and Gu]{furuta2021co}
Hiroki Furuta, Tadashi Kozuno, Tatsuya Matsushima, Yutaka Matsuo, and Shixiang~Shane Gu.
\newblock Co-adaptation of algorithmic and implementational innovations in inference-based deep reinforcement learning.
\newblock \emph{Advances in neural information processing systems}, 34:\penalty0 9828--9842, 2021.

\bibitem[Gilks et~al.(1995)Gilks, Richardson, and Spiegelhalter]{gilks1995markov}
Walter~R Gilks, Sylvia Richardson, and David Spiegelhalter.
\newblock \emph{Markov chain Monte Carlo in practice}.
\newblock CRC press, 1995.

\bibitem[Gordon et~al.(1993)Gordon, Salmond, and Smith]{gordon1993novel}
Neil~J Gordon, David~J Salmond, and Adrian~FM Smith.
\newblock Novel approach to nonlinear/non-gaussian bayesian state estimation.
\newblock In \emph{IEE proceedings F (radar and signal processing)}, volume 140, pages 107--113. IET, 1993.

\bibitem[Gorsane et~al.(2022)Gorsane, Mahjoub, de~Kock, Dubb, Singh, and Pretorius]{gorsane2022towards}
Rihab Gorsane, Omayma Mahjoub, Ruan~John de~Kock, Roland Dubb, Siddarth Singh, and Arnu Pretorius.
\newblock Towards a standardised performance evaluation protocol for cooperative marl.
\newblock \emph{Advances in Neural Information Processing Systems}, 35:\penalty0 5510--5521, 2022.

\bibitem[Grill et~al.(2020)Grill, Altch{\'e}, Tang, Hubert, Valko, Antonoglou, and Munos]{grill2020monte}
Jean-Bastien Grill, Florent Altch{\'e}, Yunhao Tang, Thomas Hubert, Michal Valko, Ioannis Antonoglou, and R{\'e}mi Munos.
\newblock Monte-carlo tree search as regularized policy optimization.
\newblock In \emph{International Conference on Machine Learning}, pages 3769--3778. PMLR, 2020.

\bibitem[Gu et~al.(2015)Gu, Ghahramani, and Turner]{gu2015neural}
Shixiang~Shane Gu, Zoubin Ghahramani, and Richard~E Turner.
\newblock Neural adaptive sequential monte carlo.
\newblock \emph{Advances in neural information processing systems}, 28, 2015.

\bibitem[Guez et~al.(2018)Guez, Mirza, Gregor, Kabra, Racaniere, Weber, Raposo, Santoro, Orseau, Eccles, Wayne, Silver, Lillicrap, and Valdes]{boxobanlevels}
Arthur Guez, Mehdi Mirza, Karol Gregor, Rishabh Kabra, Sebastien Racaniere, Theophane Weber, David Raposo, Adam Santoro, Laurent Orseau, Tom Eccles, Greg Wayne, David Silver, Timothy Lillicrap, and Victor Valdes.
\newblock An investigation of model-free planning: boxoban levels.
\newblock https://github.com/deepmind/boxoban-levels/, 2018.

\bibitem[Haarnoja et~al.(2017)Haarnoja, Tang, Abbeel, and Levine]{haarnoja2017reinforcement}
Tuomas Haarnoja, Haoran Tang, Pieter Abbeel, and Sergey Levine.
\newblock Reinforcement learning with deep energy-based policies.
\newblock In \emph{International conference on machine learning}, pages 1352--1361. PMLR, 2017.

\bibitem[Haarnoja et~al.(2018)Haarnoja, Zhou, Abbeel, and Levine]{haarnoja2018soft}
Tuomas Haarnoja, Aurick Zhou, Pieter Abbeel, and Sergey Levine.
\newblock Soft actor-critic: Off-policy maximum entropy deep reinforcement learning with a stochastic actor.
\newblock In \emph{International conference on machine learning}, pages 1861--1870. PMLR, 2018.

\bibitem[Hachiya et~al.(2009)Hachiya, Peters, and Sugiyama]{hachiya2009efficient}
Hirotaka Hachiya, Jan Peters, and Masashi Sugiyama.
\newblock Efficient sample reuse in em-based policy search.
\newblock In \emph{Machine Learning and Knowledge Discovery in Databases: European Conference, ECML PKDD 2009, Bled, Slovenia, September 7-11, 2009, Proceedings, Part I 20}, pages 469--484. Springer, 2009.

\bibitem[He et~al.(2016)He, Zhang, Ren, and Sun]{he2016deep}
Kaiming He, Xiangyu Zhang, Shaoqing Ren, and Jian Sun.
\newblock Deep residual learning for image recognition.
\newblock In \emph{Proceedings of the IEEE conference on computer vision and pattern recognition}, pages 770--778, 2016.

\bibitem[Hochreiter and Schmidhuber(1997)]{hochreiter1997long}
Sepp Hochreiter and J{\"u}rgen Schmidhuber.
\newblock Long short-term memory.
\newblock \emph{Neural computation}, 9\penalty0 (8):\penalty0 1735--1780, 1997.

\bibitem[Hubert et~al.(2021)Hubert, Schrittwieser, Antonoglou, Barekatain, Schmitt, and Silver]{hubert2021learning}
Thomas Hubert, Julian Schrittwieser, Ioannis Antonoglou, Mohammadamin Barekatain, Simon Schmitt, and David Silver.
\newblock Learning and planning in complex action spaces.
\newblock In \emph{International Conference on Machine Learning}, pages 4476--4486. PMLR, 2021.

\bibitem[Kappen et~al.(2012)Kappen, G{\'o}mez, and Opper]{kappen2012optimal}
Hilbert~J Kappen, Vicen{\c{c}} G{\'o}mez, and Manfred Opper.
\newblock Optimal control as a graphical model inference problem.
\newblock \emph{Machine learning}, 87:\penalty0 159--182, 2012.

\bibitem[Kitagawa(1996)]{kitagawa1996monte}
Genshiro Kitagawa.
\newblock Monte carlo filter and smoother for non-gaussian nonlinear state space models.
\newblock \emph{Journal of computational and graphical statistics}, 5\penalty0 (1):\penalty0 1--25, 1996.

\bibitem[Lazaric et~al.(2007)Lazaric, Restelli, and Bonarini]{lazaric2007reinforcement}
Alessandro Lazaric, Marcello Restelli, and Andrea Bonarini.
\newblock Reinforcement learning in continuous action spaces through sequential monte carlo methods.
\newblock \emph{Advances in neural information processing systems}, 20, 2007.

\bibitem[Levine(2018)]{levine2018reinforcement}
Sergey Levine.
\newblock Reinforcement learning and control as probabilistic inference: Tutorial and review.
\newblock \emph{arXiv preprint arXiv:1805.00909}, 2018.

\bibitem[Levine and Koltun(2013)]{levine2013guided}
Sergey Levine and Vladlen Koltun.
\newblock Guided policy search.
\newblock In \emph{International conference on machine learning}, pages 1--9. PMLR, 2013.

\bibitem[Levy(1992)]{levy1992stochastic}
Haim Levy.
\newblock Stochastic dominance and expected utility: Survey and analysis.
\newblock \emph{Management Science}, pages 555--593, 1992.

\bibitem[Li and Todorov(2004)]{li2004iterative}
Weiwei Li and Emanuel Todorov.
\newblock Iterative linear quadratic regulator design for nonlinear biological movement systems.
\newblock In \emph{First International Conference on Informatics in Control, Automation and Robotics}, volume~2, pages 222--229. SciTePress, 2004.

\bibitem[Li et~al.(2017)Li, Turner, and Liu]{li2017approximate}
Yingzhen Li, Richard~E Turner, and Qiang Liu.
\newblock Approximate inference with amortised mcmc.
\newblock \emph{arXiv preprint arXiv:1702.08343}, 2017.

\bibitem[Lioutas et~al.(2022)Lioutas, Lavington, Sefas, Niedoba, Liu, Zwartsenberg, Dabiri, Wood, and Scibior]{lioutas2022critic}
Vasileios Lioutas, Jonathan~Wilder Lavington, Justice Sefas, Matthew Niedoba, Yunpeng Liu, Berend Zwartsenberg, Setareh Dabiri, Frank Wood, and Adam Scibior.
\newblock Critic sequential monte carlo.
\newblock In \emph{The Eleventh International Conference on Learning Representations}, 2022.

\bibitem[Liu et~al.(2019)Liu, Chen, Yu, Zhai, Zhou, and Liu]{liu2018watch}
Anji Liu, Jianshu Chen, Mingze Yu, Yu~Zhai, Xuewen Zhou, and Ji~Liu.
\newblock Watch the unobserved: A simple approach to parallelizing monte carlo tree search.
\newblock In \emph{International Conference on Learning Representations}, 2019.

\bibitem[Liu and Chen(1998)]{liu1998sequential}
Jun~S Liu and Rong Chen.
\newblock Sequential monte carlo methods for dynamic systems.
\newblock \emph{Journal of the American statistical association}, 93\penalty0 (443):\penalty0 1032--1044, 1998.

\bibitem[Liu et~al.(2001)Liu, Chen, and Logvinenko]{liu2001theoretical}
Jun~S Liu, Rong Chen, and Tanya Logvinenko.
\newblock A theoretical framework for sequential importance sampling with resampling.
\newblock In \emph{Sequential Monte Carlo methods in practice}, pages 225--246. Springer, 2001.

\bibitem[Liu et~al.(2022)Liu, Cen, Isenbaev, Liu, Wu, Li, and Zhao]{liu2022constrained}
Zuxin Liu, Zhepeng Cen, Vladislav Isenbaev, Wei Liu, Steven Wu, Bo~Li, and Ding Zhao.
\newblock Constrained variational policy optimization for safe reinforcement learning.
\newblock In \emph{International Conference on Machine Learning}, pages 13644--13668. PMLR, 2022.

\bibitem[Mann and Whitney(1947)]{mann1947test}
Henry~B Mann and Donald~R Whitney.
\newblock On a test of whether one of two random variables is stochastically larger than the other.
\newblock \emph{The annals of mathematical statistics}, pages 50--60, 1947.

\bibitem[Maskell and Gordon(2001)]{maskell2001tutorial}
Simon Maskell and Neil Gordon.
\newblock A tutorial on particle filters for on-line nonlinear/non-gaussian bayesian tracking.
\newblock \emph{IEE Target Tracking: Algorithms and Applications (Ref. No. 2001/174)}, pages 2--1, 2001.

\bibitem[Moerland et~al.(2018)Moerland, Broekens, Plaat, and Jonker]{moerland2018a0c}
Thomas~M Moerland, Joost Broekens, Aske Plaat, and Catholijn~M Jonker.
\newblock A0c: Alpha zero in continuous action space.
\newblock \emph{arXiv preprint arXiv:1805.09613}, 2018.

\bibitem[Montgomery and Levine(2016)]{montgomery2016guided}
William Montgomery and Sergey Levine.
\newblock Guided policy search as approximate mirror descent.
\newblock \emph{arXiv preprint arXiv:1607.04614}, 2016.

\bibitem[Nair et~al.(2020)Nair, Gupta, Dalal, and Levine]{nair2020awac}
Ashvin Nair, Abhishek Gupta, Murtaza Dalal, and Sergey Levine.
\newblock Awac: Accelerating online reinforcement learning with offline datasets.
\newblock \emph{arXiv preprint arXiv:2006.09359}, 2020.

\bibitem[Neumann et~al.(2011)]{neumann2011variational}
Gerhard Neumann et~al.
\newblock Variational inference for policy search in changing situations.
\newblock In \emph{Proceedings of the 28th International Conference on Machine Learning, ICML 2011}, pages 817--824, 2011.

\bibitem[Pang et~al.(2023)Pang, Wang, Li, Chen, Xu, Zhang, and Yu]{pang2023language}
Jing-Cheng Pang, Pengyuan Wang, Kaiyuan Li, Xiong-Hui Chen, Jiacheng Xu, Zongzhang Zhang, and Yang Yu.
\newblock Language model self-improvement by reinforcement learning contemplation.
\newblock \emph{arXiv preprint arXiv:2305.14483}, 2023.

\bibitem[Peng et~al.(2019)Peng, Kumar, Zhang, and Levine]{peng2019advantage}
Xue~Bin Peng, Aviral Kumar, Grace Zhang, and Sergey Levine.
\newblock Advantage-weighted regression: Simple and scalable off-policy reinforcement learning.
\newblock \emph{arXiv preprint arXiv:1910.00177}, 2019.

\bibitem[Peters and Schaal(2007)]{peters2007reinforcement}
Jan Peters and Stefan Schaal.
\newblock Reinforcement learning by reward-weighted regression for operational space control.
\newblock In \emph{Proceedings of the 24th international conference on Machine learning}, pages 745--750, 2007.

\bibitem[Peters et~al.(2010)Peters, Mulling, and Altun]{peters2010relative}
Jan Peters, Katharina Mulling, and Yasemin Altun.
\newblock Relative entropy policy search.
\newblock In \emph{Proceedings of the AAAI Conference on Artificial Intelligence}, volume~24, pages 1607--1612, 2010.

\bibitem[Pich{\'e} et~al.(2018)Pich{\'e}, Thomas, Ibrahim, Bengio, and Pal]{piche2018probabilistic}
Alexandre Pich{\'e}, Valentin Thomas, Cyril Ibrahim, Yoshua Bengio, and Chris Pal.
\newblock Probabilistic planning with sequential monte carlo methods.
\newblock In \emph{International Conference on Learning Representations}, 2018.

\bibitem[Pitt and Shephard(2001)]{pitt2001auxiliary}
Michael~K Pitt and Neil Shephard.
\newblock Auxiliary variable based particle filters.
\newblock \emph{Sequential Monte Carlo methods in practice}, pages 273--293, 2001.

\bibitem[Puterman(2014)]{puterman2014markov}
Martin~L Puterman.
\newblock \emph{Markov decision processes: discrete stochastic dynamic programming}.
\newblock John Wiley \& Sons, 2014.

\bibitem[Schrittwieser et~al.(2020)Schrittwieser, Antonoglou, Hubert, Simonyan, Sifre, Schmitt, Guez, Lockhart, Hassabis, Graepel, et~al.]{schrittwieser2020mastering}
Julian Schrittwieser, Ioannis Antonoglou, Thomas Hubert, Karen Simonyan, Laurent Sifre, Simon Schmitt, Arthur Guez, Edward Lockhart, Demis Hassabis, Thore Graepel, et~al.
\newblock Mastering atari, go, chess and shogi by planning with a learned model.
\newblock \emph{Nature}, 588\penalty0 (7839):\penalty0 604--609, 2020.

\bibitem[Schulman(2015)]{schulman2015trust}
John Schulman.
\newblock Trust region policy optimization.
\newblock \emph{arXiv preprint arXiv:1502.05477}, 2015.

\bibitem[Schulman et~al.(2016)Schulman, Moritz, Levine, Jordan, and Abbeel]{schulman2016high}
John Schulman, Philipp Moritz, Sergey Levine, Michael~I. Jordan, and Pieter Abbeel.
\newblock High-dimensional continuous control using generalized advantage estimation.
\newblock In Yoshua Bengio and Yann LeCun, editors, \emph{4th International Conference on Learning Representations, {ICLR} 2016, San Juan, Puerto Rico, May 2-4, 2016, Conference Track Proceedings}, 2016.
\newblock URL \url{http://arxiv.org/abs/1506.02438}.

\bibitem[Schulman et~al.(2017{\natexlab{a}})Schulman, Chen, and Abbeel]{schulman2017equivalence}
John Schulman, Xi~Chen, and Pieter Abbeel.
\newblock Equivalence between policy gradients and soft q-learning.
\newblock \emph{arXiv preprint arXiv:1704.06440}, 2017{\natexlab{a}}.

\bibitem[Schulman et~al.(2017{\natexlab{b}})Schulman, Wolski, Dhariwal, Radford, and Klimov]{schulman2017proximal}
John Schulman, Filip Wolski, Prafulla Dhariwal, Alec Radford, and Oleg Klimov.
\newblock Proximal policy optimization algorithms.
\newblock \emph{arXiv preprint arXiv:1707.06347}, 2017{\natexlab{b}}.

\bibitem[Segal(2010)]{segal2010scalability}
Richard~B Segal.
\newblock On the scalability of parallel uct.
\newblock In \emph{International Conference on Computers and Games}, pages 36--47. Springer, 2010.

\bibitem[Silver et~al.(2017)Silver, Schrittwieser, Simonyan, Antonoglou, Huang, Guez, Hubert, Baker, Lai, Bolton, et~al.]{silver2017mastering}
David Silver, Julian Schrittwieser, Karen Simonyan, Ioannis Antonoglou, Aja Huang, Arthur Guez, Thomas Hubert, Lucas Baker, Matthew Lai, Adrian Bolton, et~al.
\newblock Mastering the game of go without human knowledge.
\newblock \emph{nature}, 550\penalty0 (7676):\penalty0 354--359, 2017.

\bibitem[Silver et~al.(2018)Silver, Hubert, Schrittwieser, Antonoglou, Lai, Guez, Lanctot, Sifre, Kumaran, Graepel, et~al.]{silver2018general}
David Silver, Thomas Hubert, Julian Schrittwieser, Ioannis Antonoglou, Matthew Lai, Arthur Guez, Marc Lanctot, Laurent Sifre, Dharshan Kumaran, Thore Graepel, et~al.
\newblock A general reinforcement learning algorithm that masters chess, shogi, and go through self-play.
\newblock \emph{Science}, 362\penalty0 (6419):\penalty0 1140--1144, 2018.

\bibitem[Song et~al.(2019)Song, Abdolmaleki, Springenberg, Clark, Soyer, Rae, Noury, Ahuja, Liu, Tirumala, et~al.]{song2019v}
H~Francis Song, Abbas Abdolmaleki, Jost~Tobias Springenberg, Aidan Clark, Hubert Soyer, Jack~W Rae, Seb Noury, Arun Ahuja, Siqi Liu, Dhruva Tirumala, et~al.
\newblock V-mpo: On-policy maximum a posteriori policy optimization for discrete and continuous control.
\newblock In \emph{International Conference on Learning Representations}, 2019.

\bibitem[Sun et~al.(2018)Sun, Gordon, Boots, and Bagnell]{sun2018dual}
Wen Sun, Geoffrey~J Gordon, Byron Boots, and J~Bagnell.
\newblock Dual policy iteration.
\newblock \emph{Advances in Neural Information Processing Systems}, 31, 2018.

\bibitem[Sutton and Barto(2018)]{sutton2018reinforcement}
Richard~S Sutton and Andrew~G Barto.
\newblock \emph{Reinforcement learning: An introduction}.
\newblock MIT press, 2018.

\bibitem[Todorov et~al.(2012)Todorov, Erez, and Tassa]{todorov2012mujoco}
Emanuel Todorov, Tom Erez, and Yuval Tassa.
\newblock Mujoco: A physics engine for model-based control.
\newblock In \emph{2012 IEEE/RSJ International Conference on Intelligent Robots and Systems}, pages 5026--5033. IEEE, 2012.
\newblock \doi{10.1109/IROS.2012.6386109}.

\bibitem[Toledo(2024)]{toledo2024stoix}
Edan Toledo.
\newblock Stoix: Distributed single-agent reinforcement learning end-to-end in jax, April 2024.
\newblock URL \url{https://github.com/EdanToledo/Stoix}.

\bibitem[Toledo et~al.(2023)Toledo, Midgley, Byrne, Tilbury, Macfarlane, Courtot, and Laterre]{flashbax}
Edan Toledo, Laurence Midgley, Donal Byrne, Callum~Rhys Tilbury, Matthew Macfarlane, Cyprien Courtot, and Alexandre Laterre.
\newblock Flashbax: Streamlining experience replay buffers for reinforcement learning with jax, 2023.
\newblock URL \url{https://github.com/instadeepai/flashbax/}.

\bibitem[Toussaint and Storkey(2006)]{toussaint2006probabilistic}
Marc Toussaint and Amos Storkey.
\newblock Probabilistic inference for solving discrete and continuous state markov decision processes.
\newblock In \emph{Proceedings of the 23rd international conference on Machine learning}, pages 945--952, 2006.

\bibitem[Vieillard et~al.(2020)Vieillard, Pietquin, and Geist]{vieillard2020munchausen}
Nino Vieillard, Olivier Pietquin, and Matthieu Geist.
\newblock Munchausen reinforcement learning.
\newblock \emph{Advances in Neural Information Processing Systems}, 33:\penalty0 4235--4246, 2020.

\bibitem[Wirth et~al.(2016)Wirth, F{\"u}rnkranz, and Neumann]{wirth2016model}
Christian Wirth, Johannes F{\"u}rnkranz, and Gerhard Neumann.
\newblock Model-free preference-based reinforcement learning.
\newblock In \emph{Proceedings of the AAAI Conference on Artificial Intelligence}, volume~30, 2016.

\end{thebibliography}

\medskip

{
\small


\newpage
\appendix
\addcontentsline{toc}{section}{Appendix} 
\part{Appendix} 
\parttoc 

\clearpage


\clearpage
\section{Environments}

\label{app:environments}

\subsection{Sokoban}

\subsubsection{Overview}

\begin{table}[h]
\centering
\caption{Summary of Boxoban dataset levels}
\label{tab:dataset_levels}
\begin{tabular}{@{}ll@{}}
\toprule
Difficulty Level     & Dataset Size    \\ \midrule
Unfiltered-Train     & 900k            \\
Unfiltered-Validation & 100k           \\
Unfiltered-Test      & 1k              \\
Medium               & 450k            \\
Hard                 & 50k             \\ \bottomrule
\end{tabular}
\end{table}

We use the specific instance of Sokoban outlined in \citet{boxobanlevels} illustrated in Figure \ref{fig:soku_image} with the codebase available at \url{https://github.com/instadeepai/jumanji}. The datasets employed in this study are publicly accessible at \url{https://github.com/google-deepmind/boxoban-levels}. These datasets are split into different levels of difficulty, which are categorised in Table \ref{tab:dataset_levels}.

In this research, we always train on the \textit{Unfiltered-Train} dataset. Evaluations are performed on the \textit{Hard} dataset, important for differentiating the strongest algorithms.

\begin{figure}[h]
    \centering
    \includegraphics[width=0.3\linewidth]{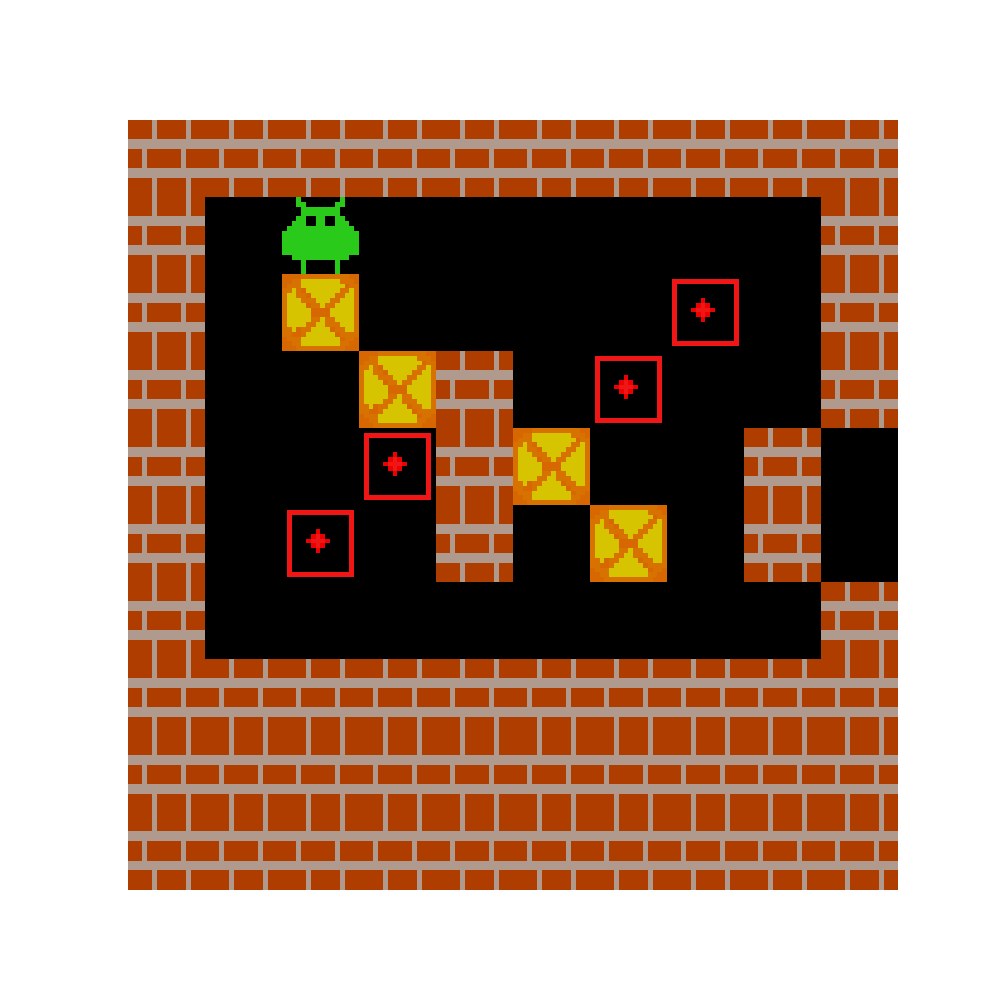} 
    \vskip -0.2in
    \caption{Example of a Boxoban Problem}
    \vskip -0.1in
    \label{fig:soku_image}
\end{figure}

\subsubsection{Network Architecture}

Observations are represented as an array of size (10,10) where each entry is an integer representing the state of a cell. We used a deep ResNet \cite{he2016deep} architecture for the torso network to embed the observation. We define a block as consisting of the following layers: [CNN, ResNet, ResNet]. Four such blocks are stacked, each characterized by specific parameters:

\begin{itemize}[noitemsep,topsep=0pt]
    \item Output Channels: (256, 256, 512, 512)
    \item Kernel Sizes: (3, 3, 3, 3)
    \item Strides: (1, 1, 1, 1)
\end{itemize}

Additionally, the architecture includes two output heads (policy and value), each comprising two layers of size 128 with ReLU activations. The output heads share the same torso network.  We use the same architecture for all algorithms.

\subsection{Rubik's Cube}

\subsubsection{Overview}

For our Rubik's Cube experiments, we utilize the implementation in \citet{bonnet2023jumanji} available at \url{https://github.com/instadeepai/jumanji}. Rubik's cube problems can be be made progressively difficult by scaling the number of random actions performed in a solution state to generate problem instances. We always perform training on a uniform distribution sampled from the range [3,7] of scrambled states, followed by exclusively evaluating on 7 scrambles. The observation is represented using an array of size (6,3,3). The action space is represented using an array of size (6,3) corresponding to each face and the depth at which it can be rotated.

\subsubsection{Network Architecture}

We utilise an embedding layer to first embed the face representations to size (6,3,3,4). The embedding is flattened and we use a torso layer consisting of a two layer residual network with layer size 512 and ReLU activations. Additionally, the architecture includes two output heads (policy and value), each comprising of a single layer of size 512 with ReLU activations.  We use the same architecture for all algorithms.

\subsection{Brax}

\subsubsection{Overview}

For experimentation on continuous settings, we make use of Brax \citep{brax2021github}. Brax is a library for rigid body simulation much like the popular MuJoCo setting \citep{todorov2012mujoco} simulator. However, Brax is written entirely in JAX \citep{jax2018github} to fully exploit the parallel processing capabilities of accelerators like GPUs and TPUs. 

It is important to note, that at the time of writing, there are 4 different physics back-ends that can be used. These back-ends have differing levels of computational complexity and the results between them are not comparable. The results generated in this paper utilise the Spring backend.

\begin{table}[h]
\centering
\caption{Observation and action space for Brax environments}
\label{tab:brax_env_spec}
\begin{tabular}{@{}lll@{}}
\toprule
Environment & Observation Size & Action Size \\ \midrule
Halfcheetah & 18      & 6       \\
Ant         & 27      & 8       \\
Humanoid    & 376     & 17      \\ \bottomrule
\end{tabular}
\end{table}

Table \ref{tab:brax_env_spec} contains the observation and action specifications of the scenarios used for our experiments. We specify the dimension size of the observation and action vectors.

\subsubsection{Network Architecture}

In practice, we found the smaller networks used in the original Brax publication to limit overall performance. We instead use a 4 layer feed forward network\footnote{except for SMC-ENTR,for which we use 3 to stabilise performance.} to represent both the value and policy network. Aside from the output layers, all layers are of size 256 and use SiLU non-linearities.  We use the same architecture for all algorithms. Unlike Rubik's Cube and Sokoban, we do not use a shared torso to learn embeddings.
\clearpage
\section{Ablations}

\label{app:ablations}

\subsection{E-step KL constraint}
\label{app:e-step-kl-constraint}

The KL constraint for the E-step within the Expectation Maximisation framework is an important component of ensuring policy improvement, due to the KL term in the ELBO. We investigate the impact of the adaptive temperature for both discrete and continuous environments. We compare SPO with an adaptive temperature updated every iteration to a variety of fixed temperatures.

\begin{figure}[h!]
\centering
\begin{minipage}[b]{0.48\textwidth}
\centering
\includegraphics[width=\textwidth]{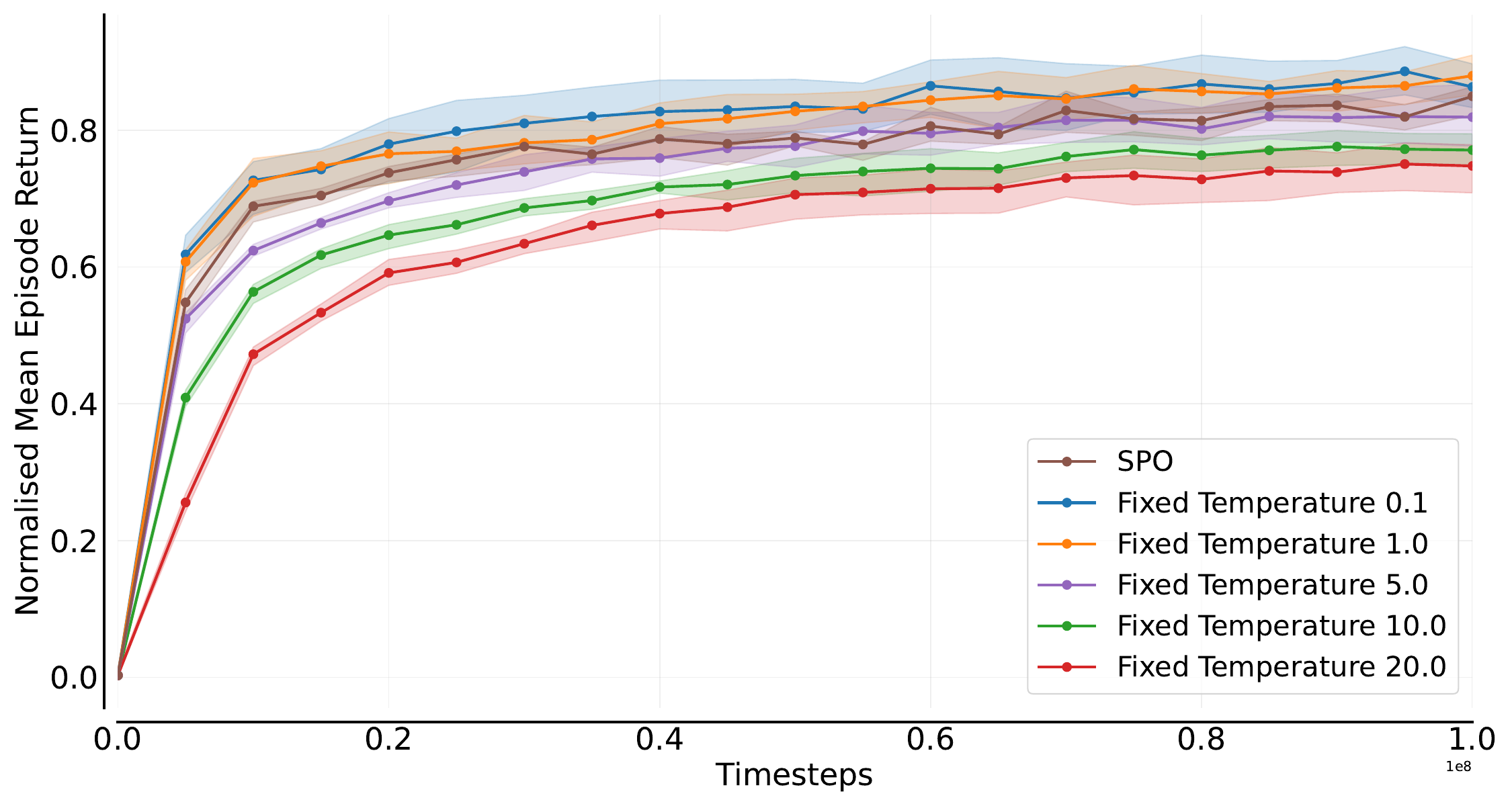}
\caption{\textbf{Brax}}
\label{fig:brax-fixed-temp}
\end{minipage}
\hfill
\begin{minipage}[b]{0.48\textwidth}
\centering
\includegraphics[width=\textwidth]{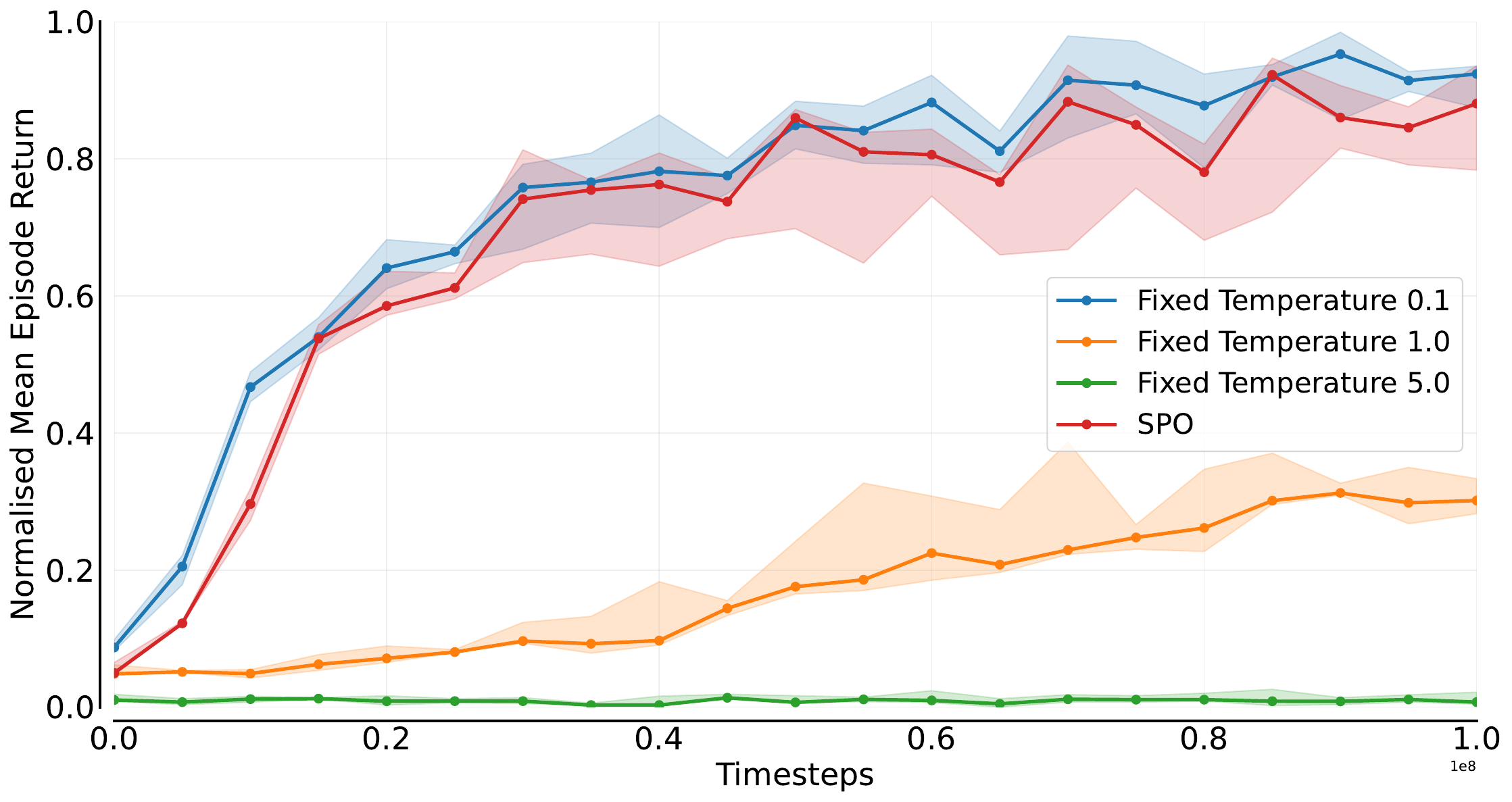}
\caption{\textbf{Sokoban and Rubik's Cube}}
\label{fig:discrete-fixed-temp}
\end{minipage}
\end{figure}

This ablation shows that SPO with an adaptive temperature is among the top performing hyperparameter settings across all environments. However we also note that it is possible to tune a temperature that works well when considering a wide range of temperatures. This is consistent with previous results in \citet{peng2019advantage} that also find practically for specific problems a fixed temperature can be used. Of course in practice having an algorithm that can learn this parameter itself is practically beneficial, removing the need for costly hyperparameter tuning, since the appropriate temperature is likely problem dependant.

Subsequently, we evaluated whether the partial optimisation of the temperature parameter $\eta$ effectively maintained the desired KL divergence constraint and how different values of this constraint affected performance.

\begin{figure}[h!]
    \centering
    \begin{minipage}[b]{0.49\textwidth}
        \centering
        \includegraphics[width=\textwidth]{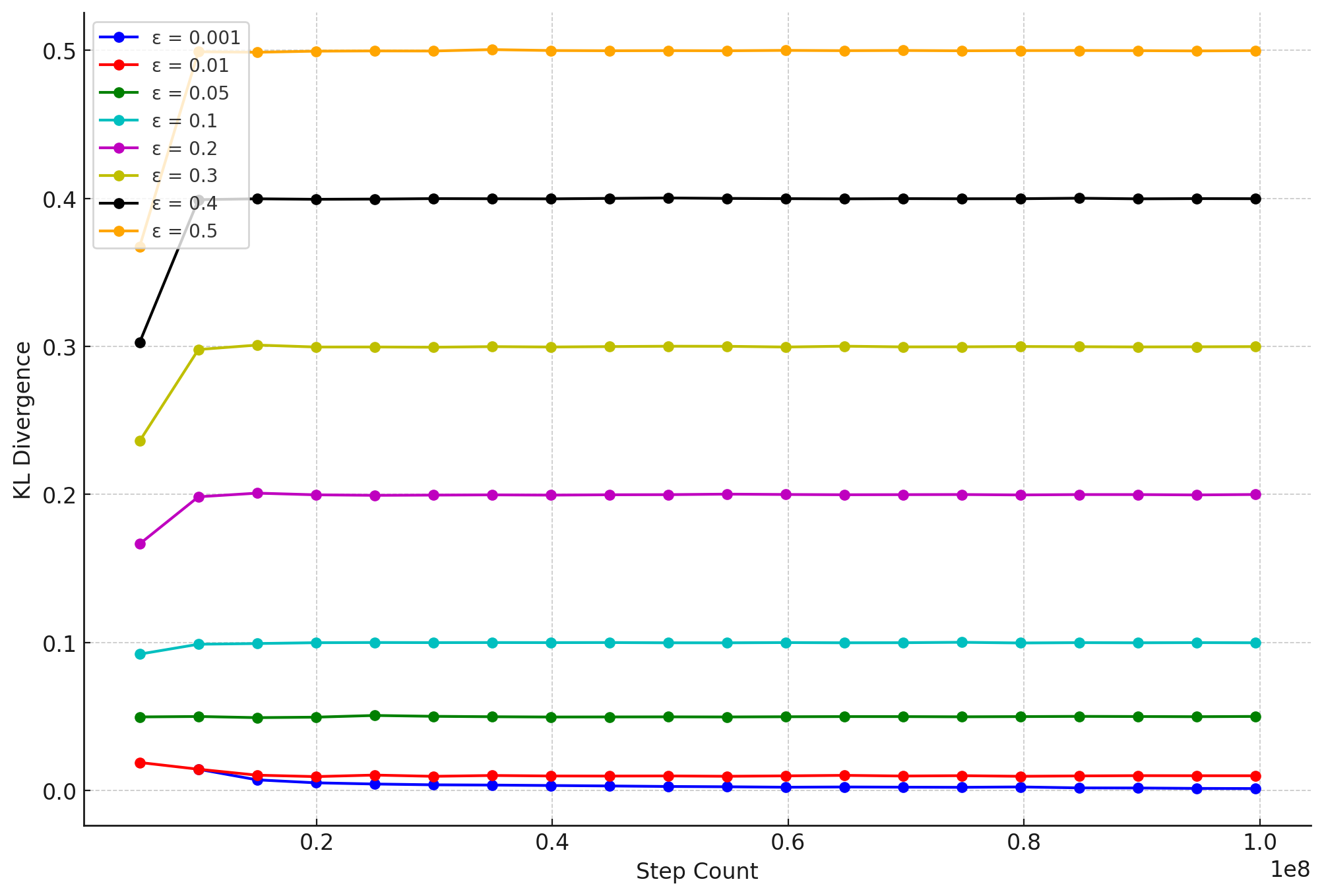}
    \end{minipage}
    \hfill
    \begin{minipage}[b]{0.49\textwidth}
        \centering
        \includegraphics[width=\textwidth]{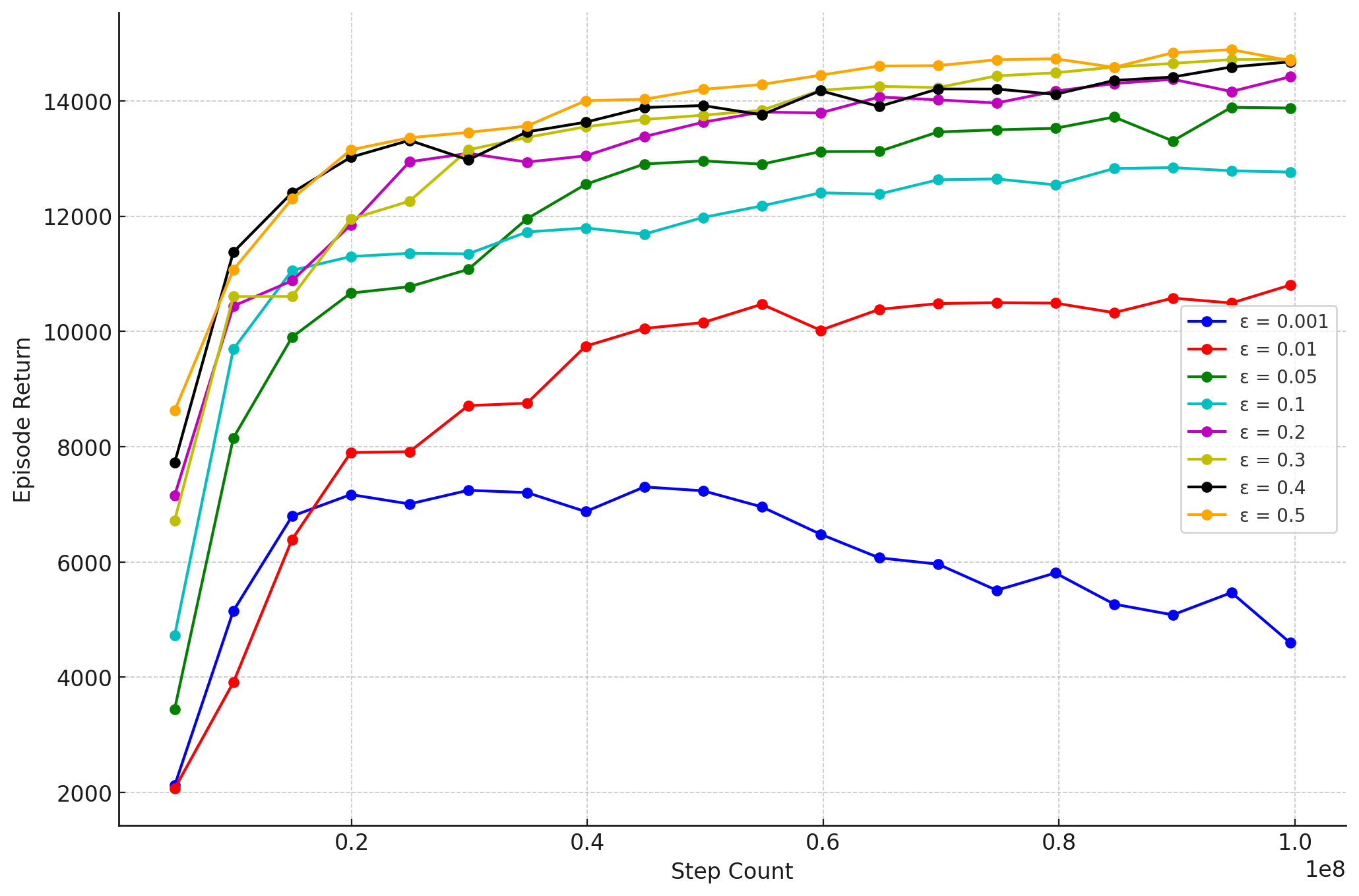}
    \end{minipage}
    \caption{(a) The estimated KL divergence between the prior policy \(\pi\) and the target policy \(q\) generated by SMC for different values of \(\epsilon\) during training on the Brax Ant task. (b) Evaluation performance during training for different values of \(\epsilon\).}
    \label{fig:kl_and_perf}
\end{figure}

As shown in Figure~\ref{fig:kl_and_perf}(a), the KL divergence constraint is tightly maintained throughout training for different values of \(\epsilon\). Figure~\ref{fig:kl_and_perf}(b) demonstrates that, for the Ant task, allowing a larger KL divergence between the prior and target policies (\(q\) and \(\pi\), respectively) leads to improved performance.

\subsection{E-step Q-value Optimisation}
\label{app:e-step-q-value}

For our analytic solution to the optimisation problem in the E-step we choose to add a value baseline such that our target distribution re-weights with respect to Advantages instead of Q-values. In practice there are no restrictions on utilising Q-values instead for the importance weight update of SMC \cref{smc:weight_update}. Below we show results on the continuous environments comparing SPO using advantages to SPO using Q-values.

\begin{figure}[h!]
    \centering
\includegraphics[width=0.6\textwidth]{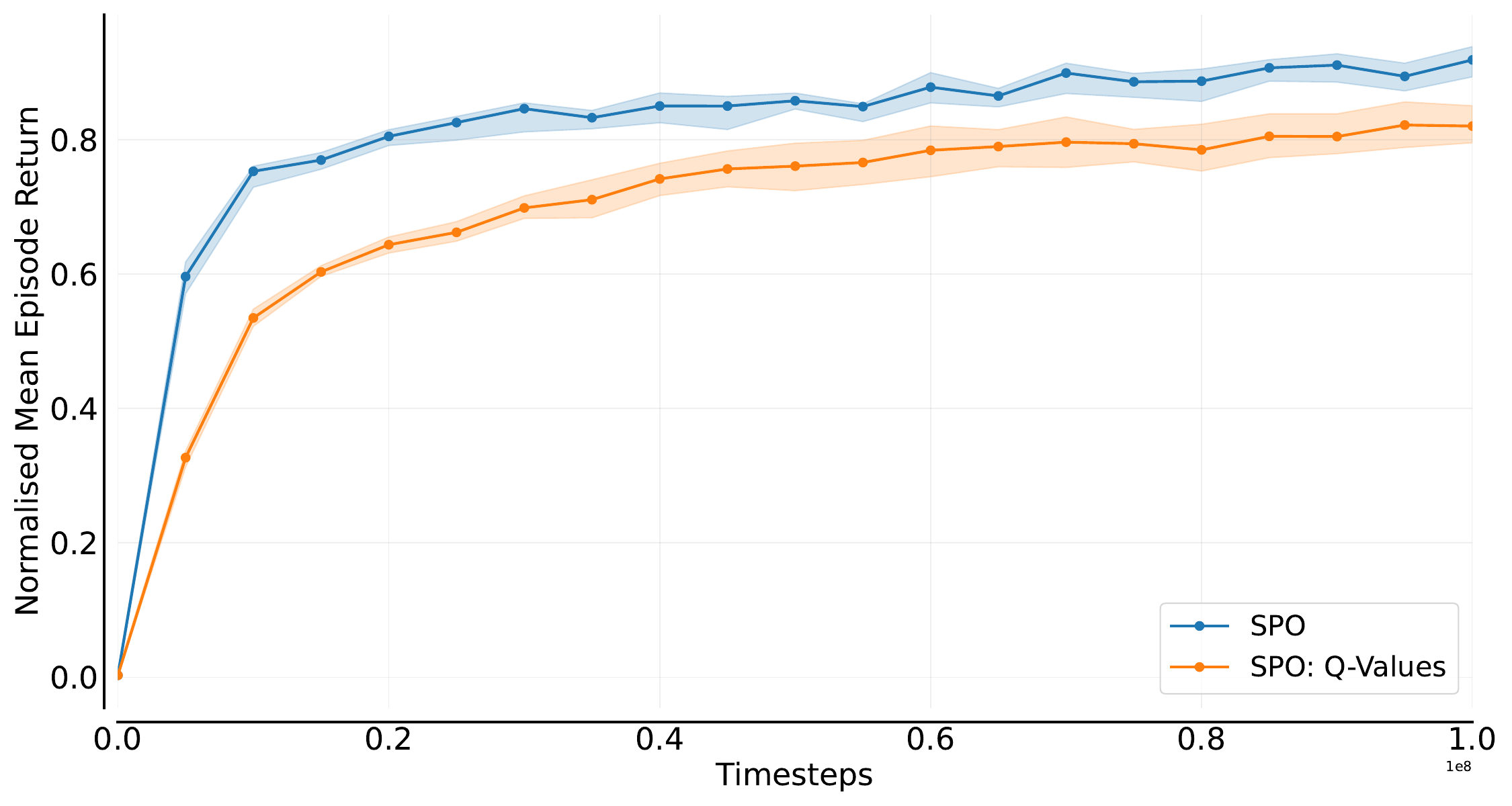}
    \caption{Q-value ablation on Brax tasks}
    \label{fig:brax_q_value}
\end{figure}

We see in Figure \ref{fig:brax_q_value}, that the use of Q values produces a reduction in performance that is relatively constant throughout training. It is possible that this reduction is due to different hyperparameters being required to effectively utilise the Q-value distribution but in order to accurately judge the effect, we kept hyperparameters constant across the runs. Secondly we note this is consistent with existing results for EM algorithms, demonstrating that Advantages outperform Q-values \citet{peng2019advantage,song2019v,nair2020awac}.

\subsection{SMC Target Estimation Validation}

Estimating the target distribution accurately is important for policy improvement in SPO. Therefore, we aim to validate the ability of SMC search to better approximate the true target distribution than a simple 1-step function evaluation as is done in algorithms such as MPO \cite{abdolmaleki2018maximum}. To do this, while the true target distribution is not directly accessible, we approximate it using an unbiased Monte Carlo oracle. We use a large computational budget of 1280 rollouts, to the end of the episode, for every state and every action. We evaluate the impact of varying planning horizons (depth) and particle counts on the KL divergence between the SMC-estimated target policy and the Monte Carlo oracle in the Sokoban environment.

\begin{figure}[h!]
    \centering
    \includegraphics[width=0.75\textwidth]{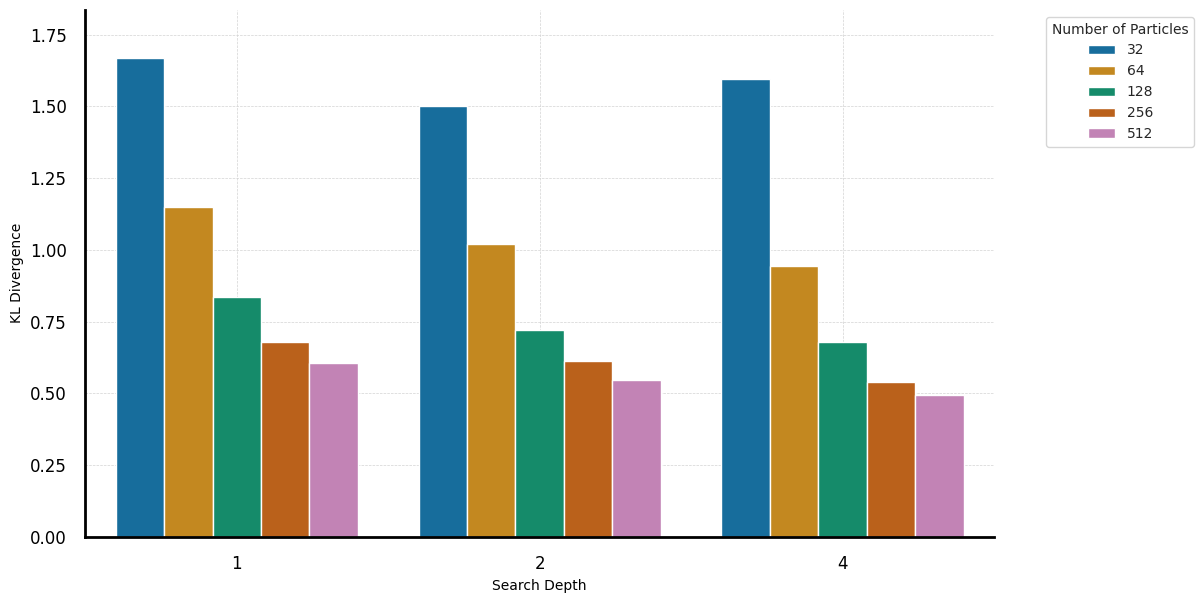}
    \caption{Comparison of the KL divergence to large Monte Carlo simulation of target policy for different planning horizons and particle counts for Sokoban}
    \label{fig:kl_oracle}
\end{figure}

As shown in Figure \ref{fig:kl_oracle}, increasing the number of particles and the planning depth reduces the KL divergence to the oracle, indicating improved estimation of the target distribution. This aligns with SMC theory~\citep{doucet2001sequential}, which suggests that more particles and deeper planning lead to better approximations. These results highlight the importance of leveraging both breadth (more particles) and depth (longer planning horizons) in SMC for accurate target estimation.
\clearpage
\section{Expanded Results}

\label{app:expanded-results}

We provide more detailed analysis of the results presented in the main body of the paper, including the final point aggregate performance of algorithms, the probability of improvement plots and the performance profiles. All plots are generated according to the methodology outlined by \citet{agarwal2021deep} and \citet{gorsane2022towards} by performing the final evaluation using parameters that achieved the best performance throughout intermediate evaluation runs during training. 

The point aggregation plots show the rankings of normalised episode returns of each algorithm when different aggregation metrics are used. The probability of improvement plots illustrate the likelihood that algorithm $X$ outperforms algorithm $Y$ on a randomly selected task. It is important to note that a statistically significant result from this metric is a probability of improvement greater than 0.5 where the confidence intervals (CIs) do not contain 0.5. The metric utilises the Mann-Whitney U-statistic \cite{mann1947test}. See \cite{agarwal2021deep} for further details. The performance profiles illustrate the fraction of the runs over all training environments from each algorithm that achieved scores higher than a specific value (given on the X-axis). Functionally, this serves the same purpose as comparing average episodic returns for each algorithm in a tabular form but in a simpler format. Additionally, if an algorithm's curve is strictly greater than or equal to another curve, this indicates "stochastic dominance" \cite{levy1992stochastic, dror2019deep}.

\subsection{Detailed Summary Results}

\begin{figure}[h!]
    \centering
    \includegraphics[width=\textwidth]{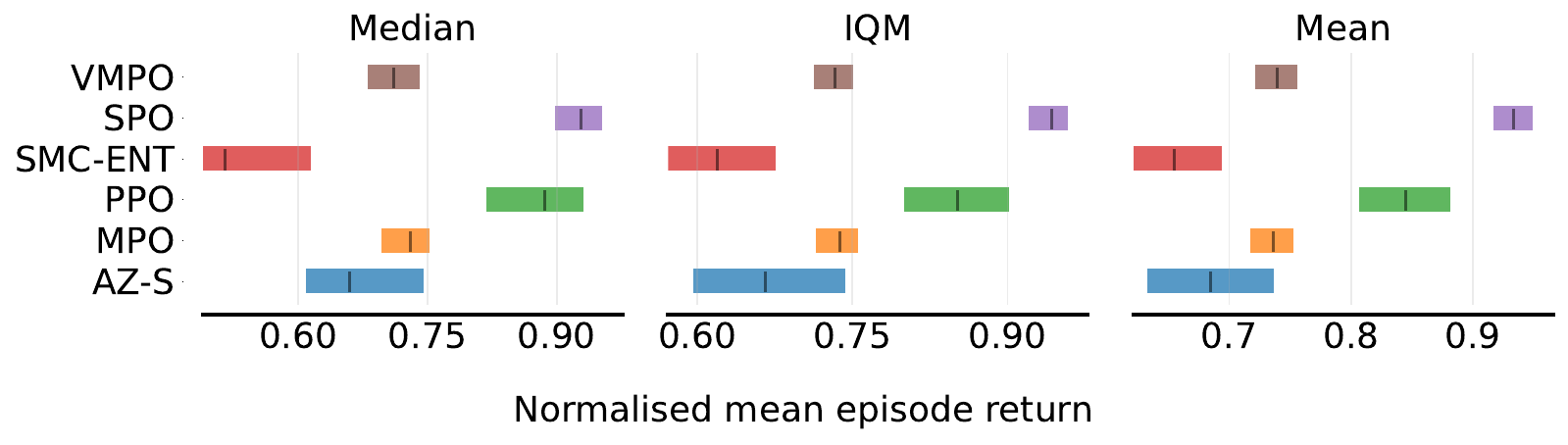}
    \caption{Aggregate point metrics for Brax suite. 95\% confidence intervals generated from stratified bootstrapping across tasks and seeds are reported.}
    \label{fig:cont-point-scores}
\end{figure}

\begin{figure}[h!]
    \centering
    \includegraphics[width=\textwidth]{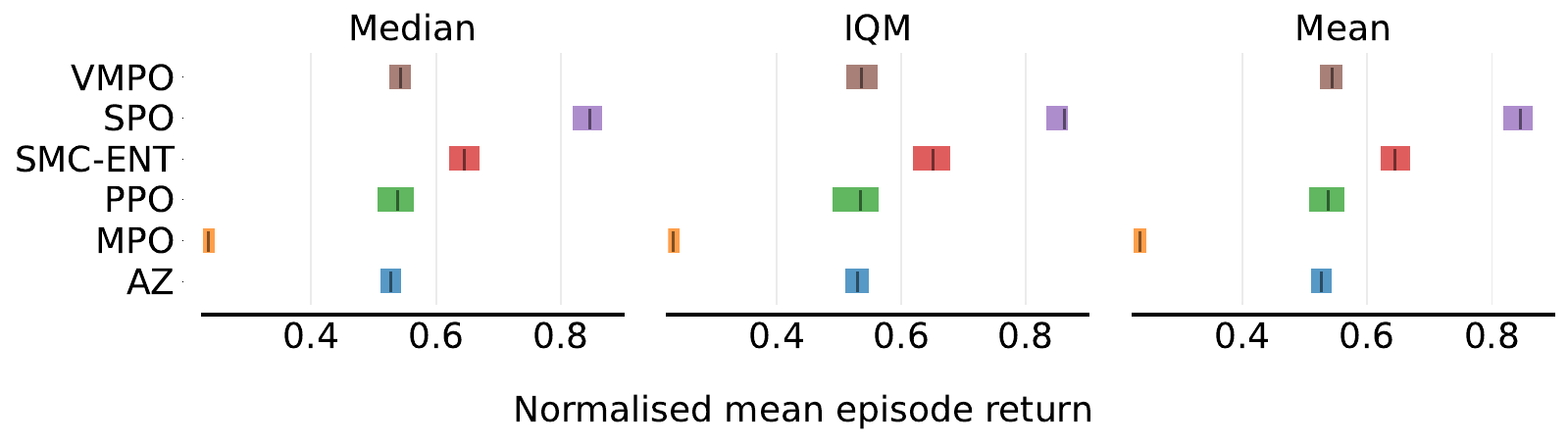}
    \caption{Aggregate point metrics for Sokoban and Rubik's Cube. 95\% confidence intervals generated from stratified bootstrapping across tasks and seeds are reported.}
    \label{fig:discrete-point-scores}
\end{figure}

In Figures \ref{fig:cont-point-scores} and \ref{fig:discrete-point-scores}, we present the absolute metrics following the recommendations by \citet{agarwal2021deep}, \citet{colas2018gep} and \citet{gorsane2022towards}. These metrics evaluate the best set of network parameters identified during 20 evenly spaced training evaluation intervals. The absolute evaluations measure performance across 10 times the number of episodes periodically assessed during training, resulting in a total of 1280 episodes evaluated. We report point estimates and their 95\% confidence intervals, which were calculated using stratified bootstrapping. \citet{agarwal2021deep} advocate for the Interquartile Mean (IQM) as a more robust and valid point estimate metric. Our results indicate that SPO surpasses all other algorithms in every point estimate. Notably, the IQM and mean point estimates demonstrate statistical significance.

\begin{figure*}[h!]
    \centering
    \begin{minipage}{0.49\textwidth}
        \centering
        \includegraphics[width=\textwidth]{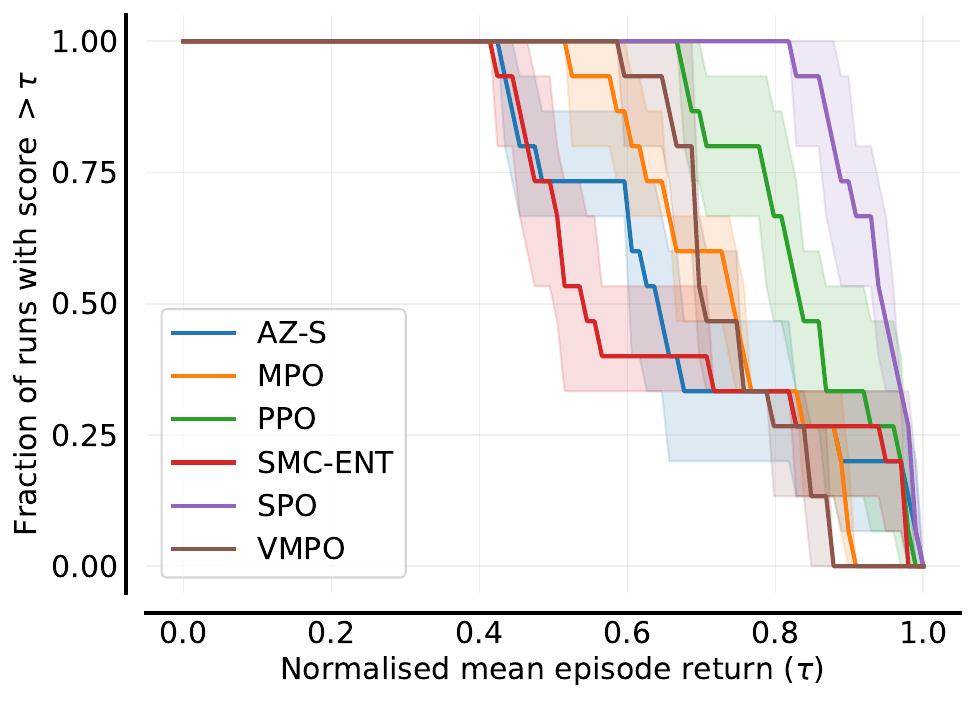}
        \textbf{(a) Brax.}
    \end{minipage}%
    \hfill
    \begin{minipage}{0.49\textwidth}
        \centering
        \includegraphics[width=\textwidth]{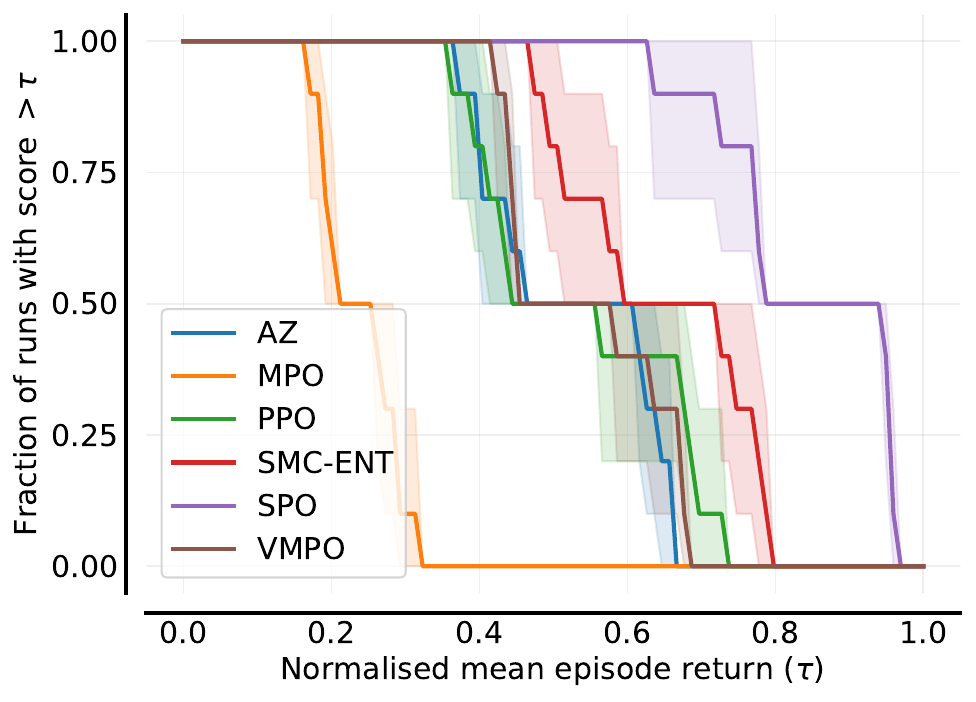}
        \textbf{(b) Sokoban and Rubik's Cube.}
    \end{minipage}
    \caption{Performance profiles. The Y-axis represents the fraction of runs that achieved greater than a specific normalised score represented on the x-axis, with shaded regions indicating 95\% confidence intervals generated from stratified bootstrapping across both tasks and random seeds.}
    \label{fig:perf-profiles}
\end{figure*}

In Figure \ref{fig:perf-profiles}, we present the performance profiles which visually illustrates the full distribution of scores across all tasks and seeds for each algorithm. We see that SPO has a higher lower bound on performance, in addition to upper bound, indicating lower variance across tasks and seeds. Additionally, we see that SPO is strictly above all other algorithms indicating that SPO is stochastically dominant\footnote{A random variable \(X\) is termed stochastically dominant over another random variable \(Y\) if \(P(X > \tau) \geq P(Y > \tau)\) for all \(\tau\), and for some \(\tau\), \(P(X > \tau) > P(Y > \tau)\).
} to all baselines.

\begin{figure*}[h!]
    \centering
    \begin{minipage}{0.49\textwidth}
        \centering
        \includegraphics[width=\textwidth]{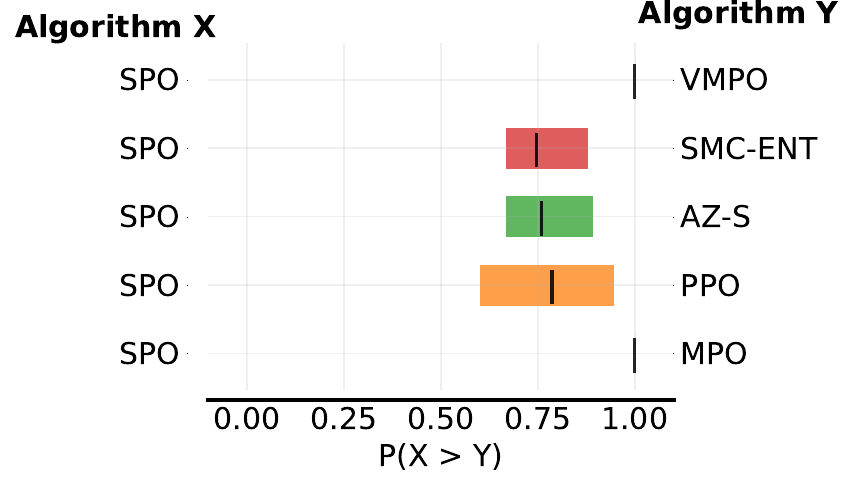}
        \textbf{(a) Brax.}
    \end{minipage}%
    \hfill
    \begin{minipage}{0.49\textwidth}
        \centering
        \includegraphics[width=\textwidth]{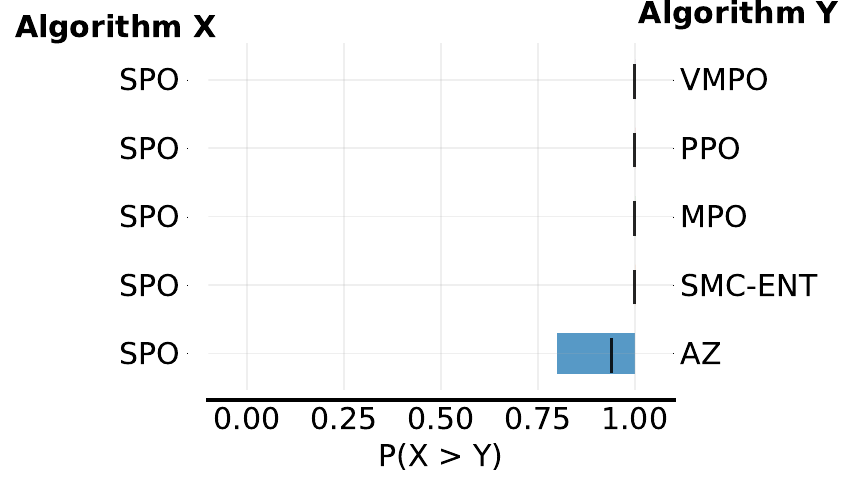}
        \textbf{(b) Sokoban and Rubik's Cube.}
    \end{minipage}
    \caption{Probability of Improvement.}
    \label{fig:prob-improvement}
\end{figure*}

Lastly, we present the probability of improvement plots in Figure \ref{fig:prob-improvement}. We see that SPO has a high probability of improvement compared to all baselines. Additionally, all probabilities are larger than 0.5 and have their CIs above 0.5 thus indicating statistical significance as specified by \citet{agarwal2021deep}. Furthermore, we see all CIs have upper bounds greater than 0.75, thereby indicating that these results are statistically meaningful as specified by \citet{bouthillier2021accounting}.

\subsection{Hardware}
\label{app:hardware}
Training was performed using a mixture of Google v4-8 and v3-8 TPUs. Each experiment was run using a single TPU and only v3-8 TPUs were used to compare wall-clock time.

\newpage

\subsection{Individual Environment Results}

All individual task results are presented in \cref{fig:all_environments}. Specifically, we present the IQM of returns achieved during the evaluation intervals throughout training.

\begin{figure*}[h!]
    \centering
    \begin{minipage}{0.49\textwidth}
        \centering
        \includegraphics[width=\textwidth]{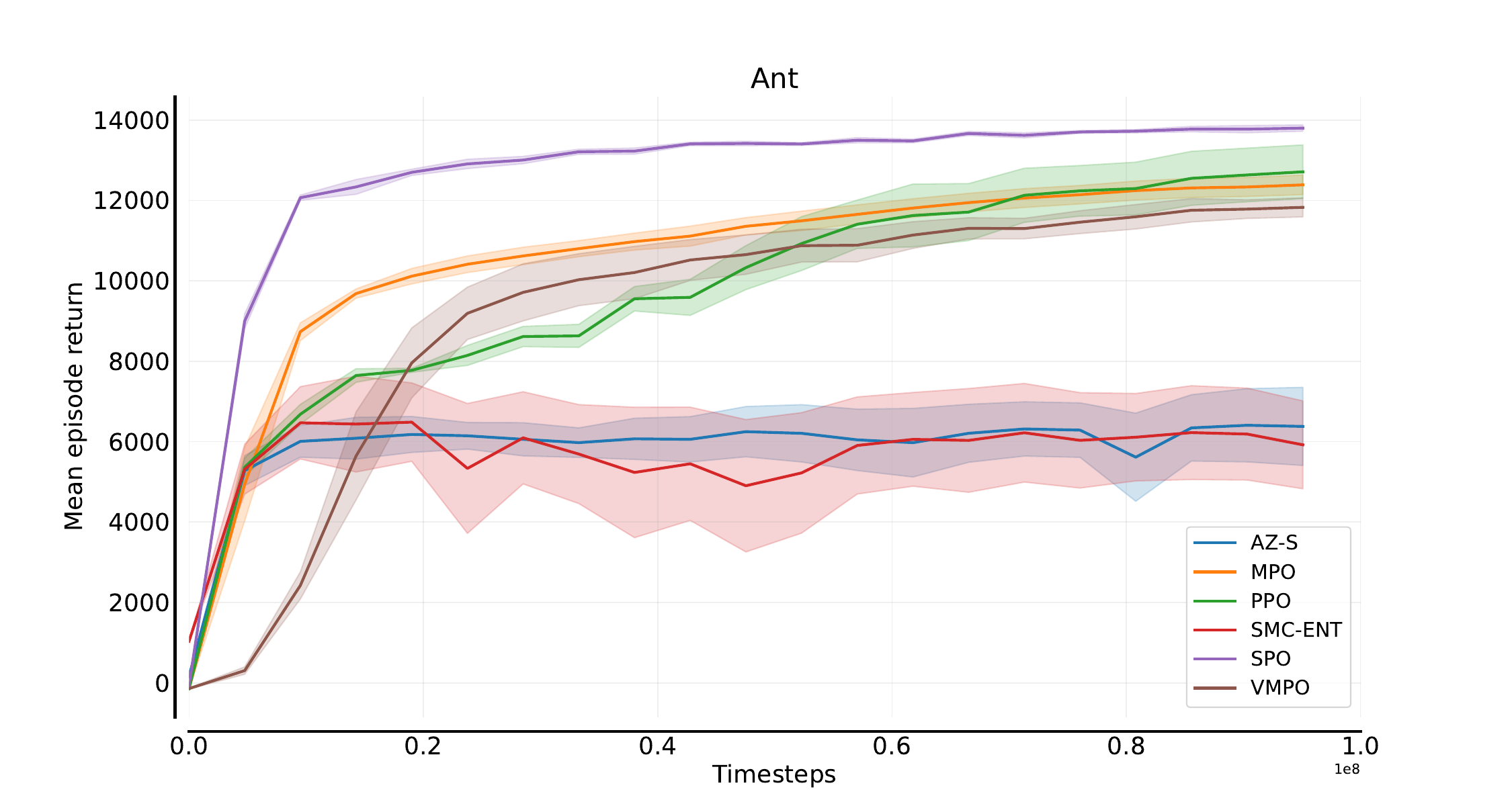}
        {(a) Ant.}
    \end{minipage}%
    \hfill
    \begin{minipage}{0.49\textwidth}
        \centering
        \includegraphics[width=\textwidth]{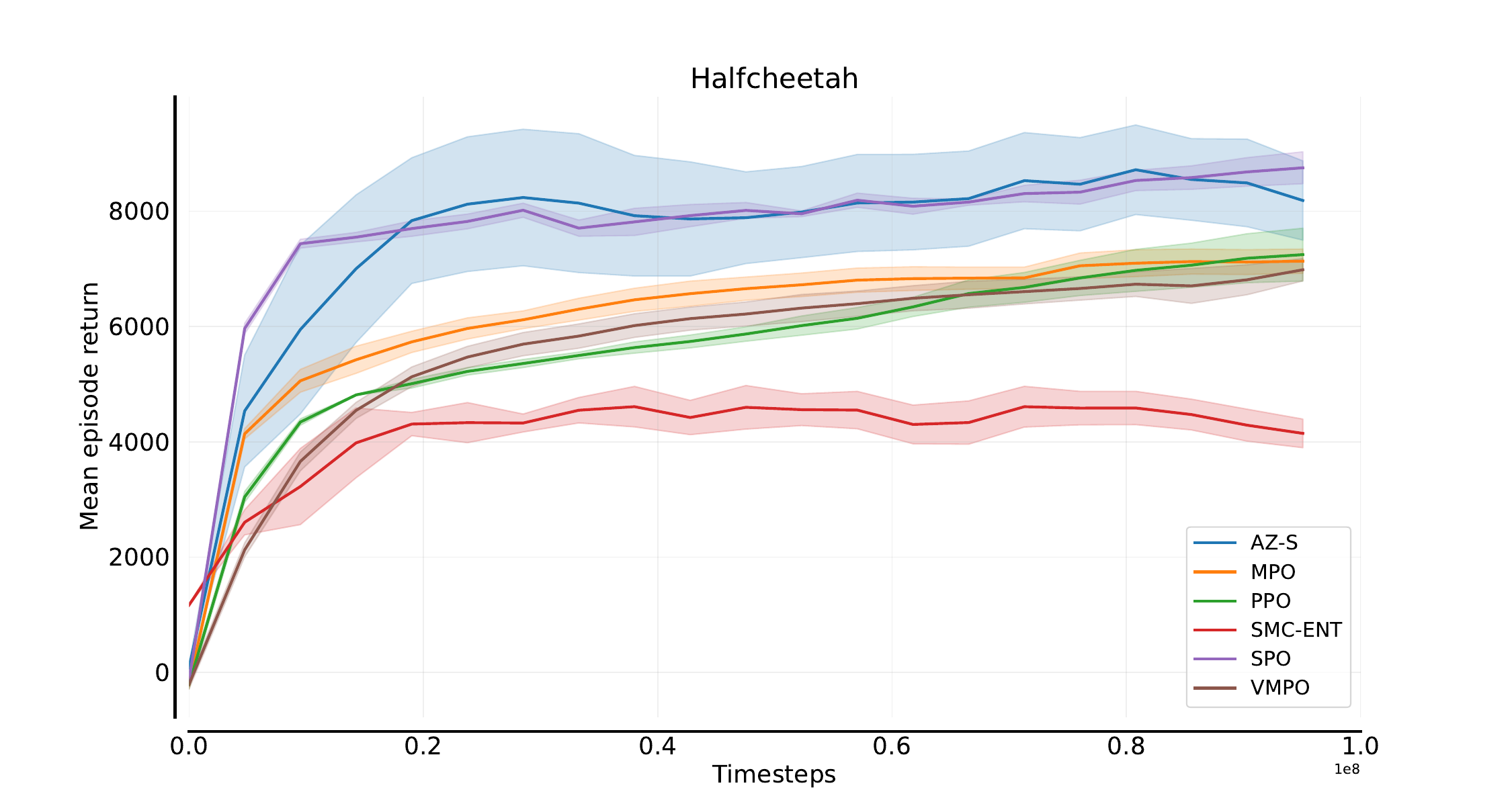}
        {(b) HalfCheetah.}
    \end{minipage}

    \vspace{0.5cm} 

    \begin{minipage}{0.49\textwidth}
        \centering
        \includegraphics[width=\textwidth]{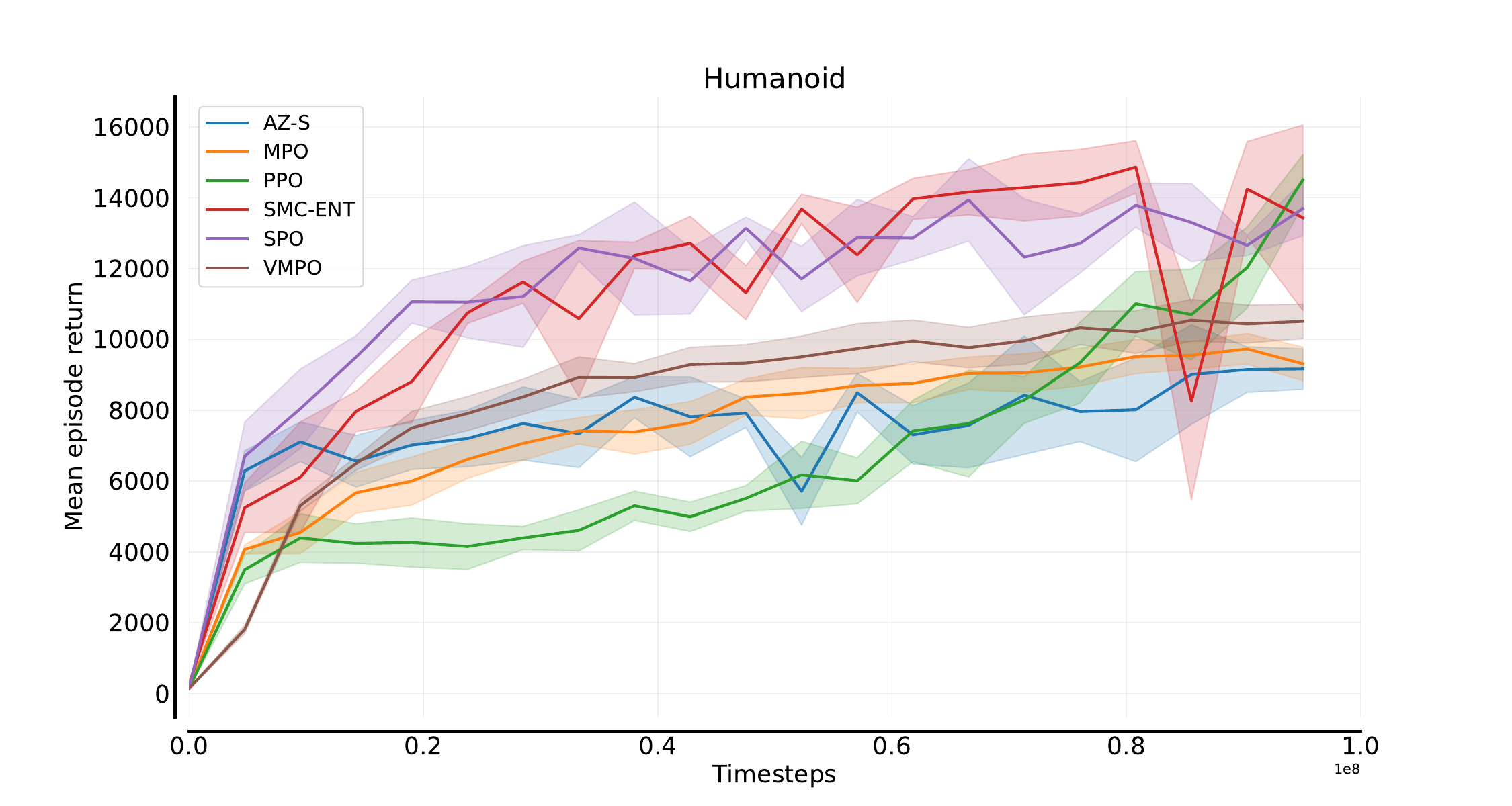}
        {(c) Humanoid.}
    \end{minipage}%
    \hfill
    \begin{minipage}{0.49\textwidth}
        \centering
        \includegraphics[width=\textwidth]{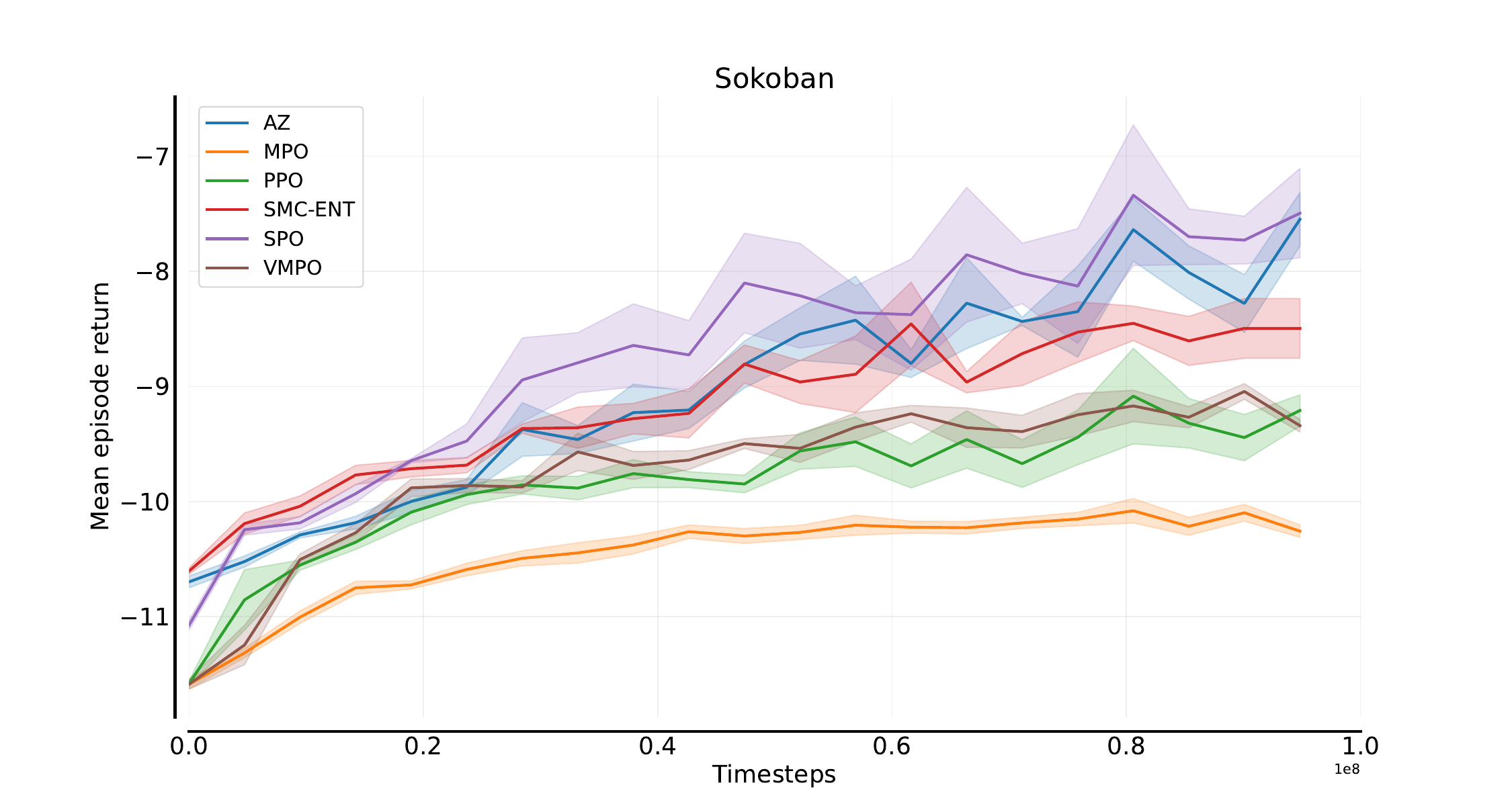}
        {(d) Sokoban.}
    \end{minipage}

    \vspace{0.5cm}

    \begin{minipage}{0.49\textwidth}
        \centering
        \includegraphics[width=\textwidth]{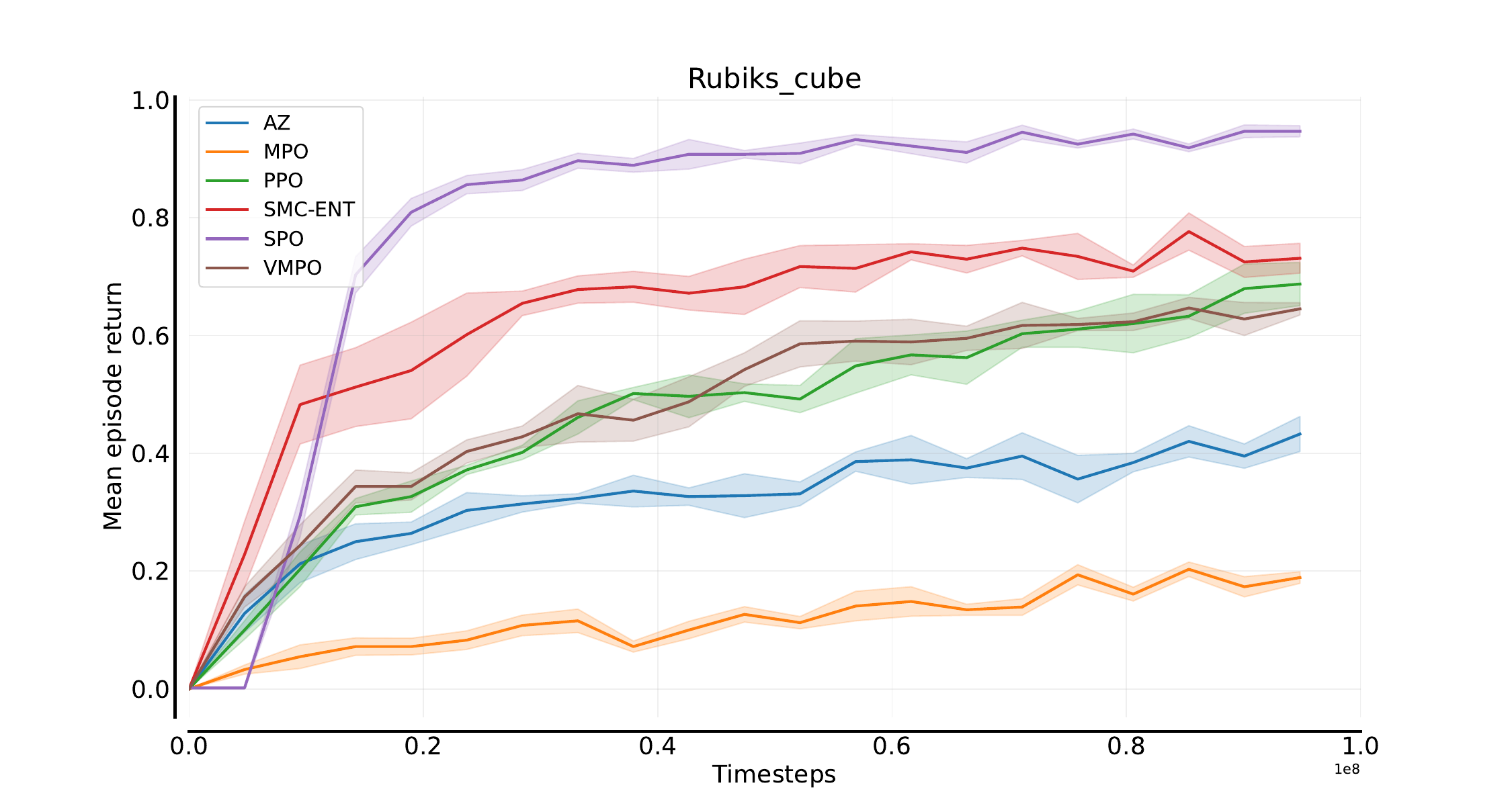}
        {(e) Rubik's Cube 7 scrambles}
    \end{minipage}%
    \hfill
    \begin{minipage}{0.49\textwidth}
        \centering
    \end{minipage}

    \caption{Performance across different environments.}
    \label{fig:all_environments}
\end{figure*}

\clearpage
\section{SPO} \label{app:SPO}

Below we outline the practical policy- based losses used within SPO,

\[\mathcal{L}_{\pi}(\theta) = -\sum_{s,a \sim{\mathcal{D}}} q_i(a|s) \log \pi_{\theta_i}(a|s)
\]

\[
    g(\eta) = \eta \varepsilon + \eta \int \mu(s) \log \left( \int \pi(a|s, \theta_i) \exp\left(\frac{A^{\bar{\pi}}(s,a)}{\eta}\right) da \right) ds.
\]

\[
\mathcal{L}_{\alpha}(\theta, \alpha) = \alpha \left( \epsilon_{\alpha} - \mathbb{E}_{s \sim p(s)} \left[ \mathrm{sg} \left[ \mathcal{D}_{\mathrm{KL}}(\pi_{\theta_{\text{old}}} \parallel \pi_{\theta}) \right] \right] \right) 
\]

\[\mathcal{L}_{KL}(\theta, \alpha) = \mathrm{sg}\left[ \alpha \right] \mathbb{E}_{s \sim p(s)} \left[ \mathcal{D}_{\mathrm{KL}}(\pi_{\theta_{\text{old}}} \parallel \pi_{\theta}) \right]\]

\subsection{Temperature Loss}

The temperature loss is calculated using advantages collected during SPO search with rollouts performed according to policy $\pi$. If resampling is utilised, after a resample the distribution over trajectories shifts towards the target distribution away from $\pi$. Therefore in order to calculate the loss $g(\eta)$, we only utilise the advantages up to the first resampling period with rollouts performed according to $\pi$. 

\subsection{Approximate Policy Iteration Algorithm}\label{app:SPO-full-algorithm}

Below we outline the overall SPO algorithm with the main loop split into the E-step and M-step.

\begin{algorithm}[h!]
\caption{SPO Algorithm}
\begin{algorithmic}[1]
\State $\mathcal{B} \leftarrow \emptyset$
\State \textbf{Initialize the policy, value function, temperature, and alpha:}
\State \hspace{1em} Initialize $\theta_0$, $\phi_0$, $\eta_0$, $\alpha_0$
\For{iteration $k = 0, 1, 2, \ldots, \text{num\_iterations}$}
    \State \textbf{Expectation Step (E-step):}
    \For{agent $m = 1, 2, \ldots, M$ \textbf{in parallel}}
        \State \hspace{1em} Initialize state $s_0$
        \For{timestep $i = 1, 2, \ldots, \text{rollout\_length}$}
            \State \hspace{1em} $\bar{a}=\{a^{(n)}\}_{n=1}^N \sim \hat{q}_{SMC}(s_i,\eta_{k},\theta_{k},\phi_{k})$ \Comment{sample-based estimate of $q_i$}
            \State \hspace{1em} $a \sim \{a^{(n)}\}_{n=1}^N$ \Comment{Sample an action to execute in the environment}
            \State \hspace{1em} $s_{i+1} \sim \mathcal{T}(s_i, a)$ \Comment{Execute action $a$, observe next state $s_{i+1}$}
            \State \hspace{1em} $r_i \sim r(s_i, a)$ \Comment{Observe reward $r_i$}
            \State \hspace{1em} Store $(s_i, r_i, \bar{a})$ in $\mathcal{B}$
        \EndFor
    \EndFor

    \State \textbf{Maximization Step (M-step):}
    \For{batch $b = 1, 2, \ldots, \text{num\_batches}$}
        \State \hspace{1em} \textbf{Sample batch from replay buffer:}
        \State \hspace{2em} Sample batch of size $N$ from replay buffer $(s, r, \bar{a}) \sim \mathcal{B}$
        \State \hspace{1em} \textbf{Update value function using GAE targets:}
        \State \hspace{2em} $\phi_{k+1} \leftarrow \arg\min_{\phi} \mathbb{E}_{s \sim \mathcal{B}} \left[ \left( V_{\text{GAE}}(s, \phi') - V(s, \phi) \right)^2 \right]$
        \State \hspace{1em} \textbf{Update parameterized policy using sample-based estimate of $q$:}
        \State \hspace{2em} $\theta_{k+1},\alpha_{k+1} \leftarrow \arg\min_{\theta} \mathbb{E}_{s \sim \mathcal{B}} \left[ \mathcal{L}_{\alpha}(\theta, \alpha) + \mathcal{L}_{KL}(\theta, \alpha)\right]$
        \State \hspace{1em} \textbf{Update dual temperature:}
        \State \hspace{2em} $\eta_{k+1} \leftarrow \arg\min_{\eta} \mathbb{E}_{s \sim \mathcal{B}} \left[g(\eta)\right]$
        \State \hspace{1em} $\phi' \leftarrow \text{polyak}(\phi', \phi_{k+1})$
    \EndFor
\EndFor
\end{algorithmic}
\end{algorithm}
\newpage

\subsection{Hyperparameters} \label{app:spo-hyperparameters}
\begin{table}[h!]
    \centering
    \caption{SPO Hyperparameters for Continuous and Discrete Environments}
    \label{tab:SPO_params}
    \small 
  \setlength{\tabcolsep}{4pt} 
    \begin{tabular}{@{}lll@{}}
        \toprule
        \textbf{Parameter} & \textbf{Continuous Environments} & \textbf{Discrete Environments} \\ \midrule
        Actor \& Critic Learning Rate  & 3e$^{-4}$ & 3e$^{-4}$ \\
        Dual Learning Rate             & 1e$^{-3}$ & 3e$^{-4}$ \\
        Discount Factor                & 0.99 & 0.99 \\
        GAE Lambda                     & 0.95 & 0.95 \\
        Replay Buffer Size             & 6.5e$^4$ & 6.5e$^4$ \\
        Batch Size                     & 32 & 64 \\
        Batch Sequence Length          & 32 & 17 \\
        Max Grad Norm                  & 0.5 & 0.5 \\
        Number of Epochs               & 128 & 16 \\
        Number of Envs                 & 1024 & 768 \\
        Rollout Length                 & 32 & 21 \\
        $\tau$ (Target Smoothing)      & 5e$^{-3}$ & 5e$^{-3}$ \\
        Number of Particles            & 16 & 16 \\
        Search Horizon                 & 4 & 4 \\
        Resample Period                & 4 & 4 \\
        Initial $\eta$                 & 10 & 0.5 \\
        Initial $\alpha$               & - & 0.5 \\
        Initial $\alpha_\mu$           & 10 & - \\
        Initial $\alpha_\Sigma$        & 500 & - \\
        $\epsilon_\eta$                & 0.2 & 0.5 \\
        $\epsilon_\alpha$              & - & 1e$^{-3}$ \\
        $\epsilon_{\alpha_\mu}$        & 5e$^{-2}$ & - \\
        $\epsilon_{\alpha_\Sigma}$     & 5e$^{-4}$ & - \\
        Dirichlet Alpha                & - & 0.03 \\
        Root Exploration Weight        & - & 0.25 \\
        \bottomrule
    \end{tabular}
\end{table}
\newpage

\subsection{Intuition for Temperature and Exploration:} The temperature parameter \(\eta\) plays a crucial role in balancing exploration and exploitation during the search process in SPO. In this work, it is derived via first choosing the desired KL divergence \(\epsilon\) between the target policy \(q_i\) generated by search and the current policy \(\pi_i\). If the temperature is too low (i.e., \(\eta\) is small), the exponential weighting \(\exp\left(A^{\bar{\pi}}(s,a)/\eta\right)\) becomes sharply peaked around the highest-advantage actions. This causes the importance weights to concentrate on a few particles, and during resampling, the particles can collapse onto a single root action. Such collapse reduces diversity in the search and renders the rest of the planning process ineffective. Conversely, if the temperature is too high (i.e., \(\eta\) is large), the weighting becomes flatter, leading to excessive exploration. While exploration is necessary, too much can prevent the search from effectively focusing on promising paths. This work shows that deriving this value via a target KL leads to stable and strong performance across different environments. 

\section{Baselines} \label{app:baselines}

For the baseline implementations, we expanded and adapted the existing implementations from Stoix\footnote{Available at: https://github.com/EdanToledo/Stoix} \cite{toledo2024stoix}. All implementations were conducted using JAX \cite{jax2018github}, with off-policy algorithms leveraging Flashbax \cite{flashbax}.

\subsection{Hyperparameters}
\label{app:baseline-hyperparameters}


\subsubsection{Continuous Control}
\begin{table}[h!]
    \centering
    \caption{Hyperparameters for PPO and Sampled AlphaZero}
    \label{tab:all_params}
    \small 
  \setlength{\tabcolsep}{4pt} 
    \begin{tabular}{@{}lll@{}}
        \toprule
        \textbf{Parameter} & \textbf{PPO} & \textbf{Sampled AlphaZero} \\ \midrule
        Actor Learning Rate        & 0.00069  & 3e$^{-4}$ \\
        Critic Learning Rate       & 0.00054  & 3e$^{-4}$ \\
        Rollout Length             & 16       & 32       \\
        Number of Epochs           & 4        & 64       \\
        Number of Minibatches      & 16       & -        \\
        Buffer Size                & -        & 65536    \\
        Batch Size                 & -        & 32       \\
        Sample Sequence Length     & -        & 32       \\
        Discount Factor            & 0.99     & 0.99     \\
        GAE Lambda                 & 0.95     & 0.95     \\
        Clip Epsilon               & 0.1      & -        \\
        Entropy Coefficient        & 0.005    & 0.005    \\
        Value Function Coefficient & 0.5      & -        \\
        Max Grad Norm              & 0.5      & 0.5      \\
        Decay Learning Rates       & True     & -        \\
        Standardize Advantages     & True     & -        \\
        Number of Simulations      & -        & 64       \\
        Dirichlet Alpha            & -        & 0.03     \\
        Dirichlet Exploration Fraction & -    & 0.25     \\
        Number of Samples          & -        & 20       \\
        Gaussian Noise Exploration Fraction & - & 0.001 \\ \bottomrule
    \end{tabular}
\end{table}

\newpage
\begin{table}[h!]
    \centering
    \caption{Hyperparameters for MPO and VMPO}
    \label{tab:all_params}
    \small 
  \setlength{\tabcolsep}{4pt} 
    \begin{tabular}{@{}lll@{}}
        \toprule
        \textbf{Parameter} & \textbf{MPO} & \textbf{VMPO} \\ \midrule
        Rollout Length             & 8        & 32       \\
        Number of Epochs           & 72       & 16       \\
        Buffer Size                & 200000   & -        \\
        Batch Size                 & 256      & -        \\
        Sample Sequence Length     & 16       & -        \\
        Period                     & 1        & -        \\
        Actor Learning Rate        & 1e$^{-4}$ & 3e$^{-4}$ \\
        Critic Learning Rate       & -        & 3e$^{-4}$ \\
        Dual Learning Rate         & 1e$^{-3}$ & 1e$^{-2}$ \\
        $\tau$ (Target Smoothing)  & 0.005    & 0.005    \\
        Discount Factor            & 0.99     & 0.99     \\
        Max Grad Norm              & 0.5      & 0.5      \\
        Decay Learning Rates       & True     & -        \\
        Number of Samples          & 128      & -        \\
        $\epsilon_\eta$                     & 0.05     & 0.05     \\
        $\epsilon_{\alpha_\mu}$               & 0.05     & 0.05     \\
        $\epsilon_{\alpha_\Sigma}$             & 0.0005   & 0.0005   \\
        Initial $\eta$        & 10.0     & 10.0     \\
        Initial $\alpha_\mu$         & 10.0     & 10.0     \\
        Initial $\alpha_\Sigma$       & 500      & 500      \\
        GAE Lambda                 & 0.95     & 0.95     \\
        Actor Target Period        & -        & 25       \\ \bottomrule
    \end{tabular}
\end{table}

\begin{table}[h!]
    \centering
    \caption{Hyperparameters for SMC-Ent}
    \label{tab:all_params}
    \small 
  \setlength{\tabcolsep}{4pt} 
    \begin{tabular}{@{}lll@{}}
        \toprule
        \textbf{Parameter} & \textbf{SMC-Ent} \\ \midrule
        Rollout Length             & 16       \\
        Number of Epochs           & 512      \\
        Buffer Size                & 500000   \\
        Batch Size                 & 512      \\
        Actor Learning Rate        & 3e$^{-4}$ \\
        Q Learning Rate            & 3e$^{-4}$ \\
        $\tau$ (Target Smoothing)  & 0.005    \\
        Discount Factor            & 0.99     \\
        Reward Scale               & 10.0     \\
        Number of Particles        & 16       \\
        Max Depth                  & 4        \\
        Resample Temperature       & 1.0      \\
        Number of Samples for Value Function & 64 \\ \bottomrule
    \end{tabular}
\end{table}

\clearpage
\subsubsection{Discrete Control}

\begin{table}[h!]
    \centering
    \caption{Hyperparameters for AlphaZero and PPO}
    \label{tab:all_params}
    \small 
  \setlength{\tabcolsep}{4pt} 
    \begin{tabular}{@{}lll@{}}
        \toprule
        \textbf{Parameter} & \textbf{AlphaZero} & \textbf{PPO} \\ \midrule
        Learning Rate              & 3e$^{-4}$ & 3e$^{-4}$ \\
        Rollout Length             & 21        & 85        \\
        Number of Epochs           & 16        & 4         \\
        Buffer Size                & 65536     & -         \\
        Batch Size                 & 64        & -         \\
        Sample Sequence Length     & 17        & -         \\
        Discount Factor            & 0.99      & 0.99      \\
        GAE Lambda                 & 0.95      & 0.95      \\
        Max Grad Norm              & 0.5       & 0.5       \\
        Number of Simulations      & 64        & -         \\
        Number of Minibatches      & -         & 64        \\
        Clip Epsilon               & -         & 0.1       \\
        Entropy Coefficient        & -         & 1e$^{-2}$ \\
        Standardize Advantages     & -         & True      \\ \bottomrule
    \end{tabular}
\end{table}

\begin{table}[H]
    \centering
    \caption{Hyperparameters for MPO and VMPO}
    \label{tab:all_params}
    \small 
  \setlength{\tabcolsep}{4pt} 
    \begin{tabular}{@{}lll@{}}
        \toprule
        \textbf{Parameter} & \textbf{MPO} & \textbf{VMPO} \\ \midrule
        Actor Learning Rate        & 2e$^{-4}$ & 3e$^{-4}$ \\
        Decay Learning Rates       & True      & -         \\
        Dual Learning Rate         & 0.02      & 0.01      \\
        Number of Epochs           & 72        & 4         \\
        $\epsilon_\eta$            & 0.07      & 0.5       \\
        $\epsilon_\alpha$          & 0.00015   & 0.001     \\
        GAE Lambda                 & 0.95      & 0.95      \\
        Discount Factor            & 0.95      & 0.99      \\
        Initial $\eta$             & 3.0       & 3.0       \\
        Initial $\alpha$           & 3.0       & 3.0       \\
        Max Grad Norm              & 0.5       & 0.5       \\
        Number of Samples          & 128       & -         \\
        Q Learning Rate            & 0.001     & -         \\
        Rollout Length             & 16        & 86        \\
        Sample Sequence Length     & 17        & -         \\
        $\tau$ (Target Smoothing)  & 0.005     & -         \\
        Batch Size                 & 256       & -         \\
        Buffer Size                & 500000    & -         \\
        Actor Target Period        & -         & 64        \\ \bottomrule
    \end{tabular}
\end{table}
\newpage

\begin{table}[H]
    \centering
    \caption{Hyperparameters for SMC-Ent}
    \label{tab:all_params}
    \small 
  \setlength{\tabcolsep}{4pt} 
    \begin{tabular}{@{}lll@{}}
        \toprule
        \textbf{Parameter} & \textbf{SMC-Ent} \\ \midrule
        Rollout Length             & 32        \\
        Number of Epochs           & 64        \\
        Buffer Size                & 1000000   \\
        Batch Size                 & 2048      \\
        Q Learning Rate            & 3e$^{-4}$ \\
        $\tau$ (Target Smoothing)  & 0.005     \\
        Discount Factor            & 0.97      \\
        Max Grad Norm              & 0.5       \\
        Huber Loss Parameter       & 1.0       \\
        Entropy Temperature        & 0.03      \\
        Munchausen Coefficient     & 0.9       \\
        Clip Value Min             & -1e$^{3}$ \\
        Number of Particles        & 16        \\
        Search Horizon             & 4         \\
        Resampling Period          & 4         \\
        Resample Temperature     & 0.1       \\
        Reward Scaling             & 10.0 (Rubik's Cube) / 1.0 (Sokoban) \\ \bottomrule
    \end{tabular}
\end{table}
\clearpage
\section{Expectation Maximisation}

\label{app:em}

\subsection{Overview of methods}
\label{app:em-summary_of_em_methods}

EM algorithms differ on a small number of dimensions through which the algorithm can be understood. The first dimension, $\mathcal{G}$, is the optimization objective that is maximized in the E-step. This includes whether advantages, Q-values, or rewards are used, and with respect to which policy. We then consider whether a trust region constraint is used both in the E-step and M-step.

With a defined optimization objective, various methods can be used to estimate it. Most methods derive an analytic solution to the optimization problem and then estimate this distribution using techniques such as TD(0) or function approximation. For example, AlphaZero does not explicitly derive the analytic solution but estimates the solution to this optimization objective through Monte Carlo Tree Search (MCTS).

\Cref{tab:em-algorithms} provides a summary of some common EM methods, highlighting their core differences.

\begin{table}[H]
  \caption{Summary of EM based Algorithms}
  \label{tab:method_summary}
  \small 
  \setlength{\tabcolsep}{4pt} 
  \centering
  \begin{tabular}{lcccccc}
    \toprule
    \textbf{Method} & \(\mathcal{G}\) & \textbf{E-step TR} & \textbf{M-step TR} & \(\mathcal{G}\) estimate & \textbf{Depth} & \textbf{Breadth} \\
    \midrule
    MPO & \(Q^{\pi_p}\) & Yes & Yes & Analytic + Function Approximation & 0 & M \\
    V-MPO & \(A^{\pi_p}\) & Yes & Yes & Analytic + n-step TD + top-k Adv & $T$ & 1 \\
    AWR & \(A^{\pi_p}\) & No & No & Analytic + n step TD & T & 1 \\
    AWAC & \(A^{\pi_p}\) & No & No & Analytic + Function Approximation  & 0 & 1 \\
    PoWER & \(\eta \log Q^{\pi_p}\) & Yes & No & Analytic + n step TD & T & 1 \\
    RWR & \(\eta \log r\) & No & No & - & 1 & 1 \\
    REPS & \(A^{\pi_p}\) & Yes & No & Analytic + TD(0) & 1 & 1 \\
    AlphaZero & \(Q^{\pi_q}\) & Yes & No & MCTS & $>0$ & $>0$ \\
    SPO & \(A^{\pi_q}\) & Yes & Yes & Analytic + SMC & $D$ & M \\
    \bottomrule
  \end{tabular}
  \label{tab:em-algorithms}
\end{table}

Note that $T$ refers to an episode length. In additon we add two important dimensions of the $\mathcal{G}$ estimate (breadth and depth). Methods like MPO estimate the analytic solution directly using a Q-function. This can be leveraged to estimate the target distribution for a selection of $M$ actions, but without leveraging depth, so rewards and future states are not used to improve the estimates. In contrast, V-MPO leverages depth to form an n-step TD estimate, but only for one of the actions, leading to a relatively poor estimate of the analytic solution.

\clearpage
\section{Proofs and Discussions}

\label{app:proof}

\subsection{E-step Analytic solution}
\label{app:em-analytic-derivation}

We outline the analytic solution to \cref{eq:constrained_optimisation}, by writing a Lagrangian equation and solving for \(q\). We can first represent our constrained optimisation exactly with the following optimisation problem and associated 2 constraints. We add a state dependent baseline $V(s)$ to the optimisation objective and notate \(Q^{q}(s, a) - V^{q}(s)\) as \(A^{q}(s,a)\).

\begin{equation}
\begin{aligned}
\max_q & \, \int \mu_q(s) \left[ \int q(a|s) \left[ Q^{q}(s, a) - V^{q}(s) \right] da \right] ds \\
\text{s.t.} & \, \int \mu_q(s) \left[ \text{KL}(q(a|s) \| \pi(a|s, \theta_i)) \right] ds < \epsilon, \\
& \, \int \mu_q(s) \left[ \int q(a|s) da \right] ds = 1.
\end{aligned}
\end{equation}

The following Lagrangian \(L\) can be constructed,

\begin{equation}
\begin{aligned}
L(q, \eta, \gamma) = & \, \int \mu_q(s) \left[ \int q(a|s) A^{q}(s, a) \, da \right] \, ds \\
& + \eta \left( \epsilon - \int \mu_q(s) \left[ \int q(a|s) \log \left( \frac{q(a|s)}{\pi(a|s, \theta_i)} \right) \, da \right] \, ds \right) \\
& + \gamma \left( 1 - \int \mu_q(s) \left[ \int q(a|s) \, da \right] \, ds \right).
\end{aligned}
\end{equation}

We can then solve for the value of \(q\) that maximises this expression by taking the derivative with respect to \(q\) and setting it to \(0\).

\begin{equation}
\begin{aligned}
\frac{\partial L(q, \eta, \gamma)}{\partial q} &= A^{q}(s, a) - \eta \log q(a|s) + \eta \log \pi(a|s, \theta_i) - (\eta + \gamma),
\end{aligned}
\end{equation}

The analytic form for the optimal distribution \(q\) can then be calculated as:

\begin{equation}
\begin{aligned}
q(a|s) &= \pi(a|s, \theta_i) \exp \left( \frac{A^{q}(s, a)}{\eta} \right) \exp \left( -\frac{\eta + \gamma}{\eta} \right).
\end{aligned}
\end{equation}

\subsection{E-step KL constraining temperature}
\label{app:em-kl-constraining-temp}

We now derive the dual function to be minimised in order to obtain $\eta$ within the analytic solution for $q$, which is crucial for enforcing the KL constraint.

The final term $-\frac{\eta + \gamma}{\eta} $ acts as a normalising constant since it is independent of $q$ and therefore we can construct the following equality.

\begin{equation}
\begin{aligned}
\exp \left( \frac{\eta + \gamma}{\eta} \right) = \int \pi(a|s, \theta_i) \exp \left( \frac{A^{q}(a, s)}{\eta} \right) da,
\end{aligned}
\end{equation}

Now that we have expressions for $-\frac{\eta + \gamma}{\eta} $ and $q$ we can form the dual function $g(\eta)$ by substituting these terms back into the original Lagrangian. After simplifying we recover the following expression.

\begin{equation}
\begin{aligned}
g(\eta) = & \, \eta \epsilon + \eta \int \mu_q(s) \log \left( \int \pi(a|s, \theta_i) \exp \left( \frac{A^{q}(a, s)}{\eta} \right) da \right) ds
\end{aligned}
\end{equation}

The optimal dual variable can be calculated as follows

\begin{equation}
\eta^*= \arg \min_{\eta} g(\eta).
\end{equation}

\subsection{E step: Constraint Optimisation}
\label{app:constr_scaling}

In optimising \cref{eq:eq4}, the first term $\mathbb{E}_{q(a|s)} \left[Q^q(s,a)\right]$ is dependent on the scale of reward in the environment. This can make it hard to choose $\alpha$, as the first and second terms are on arbitrary scales. Practically this can mean that for each new environment a new $\alpha$ should be used requiring costly hyperparameter tuning to find. As explored in previous work \cite{abdolmaleki2018maximum, song2019v} we choose to enforce a hard constraint, as opposed to a soft constraint. Instead of choosing $\alpha$ we choose $\epsilon$ which is the maximum KL divergence between $q$ and $\pi$ which is practically less sensitive to reward scale.  We found that choosing an $\epsilon$ to be far easier, resulting in a value that generalised across all environments explored. This is important as there has been a trend in recent years to add hyperparameters to Reinforcement Learning algorithms \cite{adkinsmethod} resulting in the need for costly hyperparameter sweeps for algorithms to work on each new problem they are applied to.

\subsection{ Proof of Proposition 1: Monotone Improvement Guarantee}

\label{app:em-monotone-improvement}

The optimality of policy $\pi_{\theta}$, $\log p_{\pi_{\theta}}(\mathcal{O} = 1)$, is lower bounded by the following ELBO objective:
\[
\mathcal{J}(q, \pi_{\theta}) = \mathbb{E}_{\tau \sim q} \left[ \sum_{t=0}^{\infty} \left( \gamma^t r_t - \alpha D_{KL}(q(\cdot | s_t) \| \pi(\cdot | s_t, \theta)) \right) \right] + \log p(\theta).
\]

We improve $\pi_{\theta}$ by optimizing the ELBO alternatively via EM. We will outline the proof of the monotonic improvement guarantee of the ELBO using the EM procedure, at the $i$-th iteration of training. We note that this holds under the assumption that we calculate the true value of the closed form solution $q_i$ in the E-step, which is in practice unlikely to be the case.

\textbf{E-step:} By the definition of E-step, we improve the ELBO with respect to $q$. We show the E-step update will increase ELBO:
\[
\begin{aligned}
q_{i+1} &= \arg \max_{q } \mathbb{E}_{q} \left[ \mathbb{E}_{a \sim q(\cdot | s)} \left[ A^{q_{i}}(s, a) \right] - \alpha D_{KL}[q(\cdot | s) \| \pi(\cdot | s, \theta_i)] \right] \\
&= \arg \max_{q} \mathbb{E}_{\tau \sim q} \left[ \sum_{t=0}^{\infty} \left( \gamma^t r_t - \alpha D_{KL}(q(\cdot | s_t) \| \pi(\cdot | s_t, \theta_i)) \right) \right] \\
&= \arg \max_{q} \mathcal{J}(q, \theta_i) \\
&\Rightarrow \mathcal{J}(q_{i+1}, \pi_{\theta_i}) \geq \mathcal{J}(q_{i}, \pi_{\theta_i}).
\end{aligned}
\]

\textbf{M-step:} We update $\theta$ by
\[
\begin{aligned}
\theta_{i+1} &= \arg \max_{\theta} \mathbb{E}_{{q_{i+1}}} \left[ \alpha \mathbb{E}_{a \sim q_{i+1}(\cdot | s)} \left[ \log \pi_{\theta}(a | s) \right] + \log p(\theta) \right] \\
&= \arg \max_{\theta} \mathbb{E}_{{q_{i+1}}} \left[ - \alpha D_{KL}[q_{i+1}(\cdot | s_t) \| \pi(\cdot | s_t, \theta)] + \log p(\theta) \right] \\
&= \arg \max_{\theta} \mathbb{E}_{{q_{i+1}}} \left[ \mathbb{E}_{a \sim q_{i+1}(\cdot | s)} \left[ A^{q_{i}}(s, a) \right] - \alpha D_{KL}[q_{i+1}(\cdot | s_t) \| \pi(\cdot | s_t, \theta)] + \log p(\theta) \right] \\
&= \arg \max_{\theta} \mathcal{J}(q_{i+1}, \pi_{\theta}).
\end{aligned}
\]

Therefore, we have: \(\mathcal{J}(q_{i+1}, \pi_{\theta_{i+1}}) \geq \mathcal{J}(q_{i+1}, \pi_{\theta_i})\). Combining these two results we have that after successive applications of the E-step and M-step,
\[
\mathcal{J}(q_{i+1}, \pi_{\theta_{i+1}}) \geq \mathcal{J}(q_{i}, \pi_{\theta_i}).
\]

\subsection{M-step objective}

\label{app:fisher}

In Reinforcement Learning it can be beneficial to constrain policies from moving too far from the current policy, often leading to increased stability or performance of an algorithm \cite{schulman2015trust, schulman2017proximal}. In this work we utilise a Gaussian prior around $\theta_i$, our current value of $\theta$, optimised from the previous iteration. 

Therefore, $\theta \sim \mathcal{N}(\mu, \Sigma)$ where  $\mu = \theta_i$ and $\Sigma^{-1} = \lambda F(\theta_i)$, where $F$ is the Fisher information matrix. The prior term in \cref{eq:m-step-pre-step} can then be written as $\log p(\theta) = -\lambda(\theta - \theta_i)^T F(\theta_i)^{-1}(\theta - \theta_i) + c$. The first term in this expression is the quadratic approximation of the KL divergence \cite{schulman2015trust} while the second term $c$ is a term that does not depend on $\theta$ and therefore when optimising \cref{eq:m-step-pre-step} with respect to $\theta$, can be dropped. Using this approximation we rewrite \cref{eq:m-step-pre-step} as:
$$
\max_{\theta} \mathbb{E}_{s \sim \mu_q(s)} \left[ \mathbb{E}_{a \sim q(a|s)} \left[ \log \pi(a|s, \theta) \right] - \lambda \, \text{KL} \left( \pi(a|s, \theta_i) \parallel \pi(a|s, \theta) \right) \right]
$$

Like the E-step, choosing $\lambda$ can be non-trivial so we convert it into a hard constraint optimisation \cref{eq:m-step}. Note that $\epsilon$ in \cref{eq:m-step} is different from \cref{eq:constrained_optimisation}. We note that considering a Gaussian prior in the M-step is not strictly required for SPO performance, however practically it adds stability to training.

\subsection{Connection to Mirror Descent Guided Policy Search}
\label{app:md-gps}

We highlight the connection between the use of Expectation Maximisation to improve the evidence lower bound and Mirror Descent Guided Policy Search (MD-GPS). Mirror Descent Guided Policy Search builds upon previous guided policy search work \citep{levine2013guided}. Rather than only enforcing the constraint between $q$ and $\pi$ at convergence (referred to a the local policy:$p_{i}$ and global policy: $\pi_{\theta}$ respectively), they constrain $q_{i}$ against the current policy $\pi_i$ at every iteration. This results in a very similar algorithm to EM optimisation. Below we provide the outline of MD-GPS algorithm, clearly showing the equivalence to EM \citep{montgomery2016guided}:

\begin{algorithm}
\caption{Mirror descent guided policy search (MDGPS): convex linear variant}
\begin{algorithmic}[1]
\For{iteration $k \in \{1, \ldots, K\}$}
    \State \textbf{C-step:} $p_i \leftarrow \arg \min_{p_i} \mathbb{E}_{p_i(\tau)}\left[\sum_{t=1}^{T} \ell(\mathbf{x}_t, \mathbf{u}_t)\right]$ such that $D_{\text{KL}}(p_i(\tau) \parallel \pi_\theta(\tau)) \leq \epsilon$
    \State \textbf{S-step:} $\pi_\theta \leftarrow \arg \min_\theta \sum_i D_{\text{KL}}(p_i(\tau) \parallel \pi_\theta(\tau))$
\EndFor
\end{algorithmic}
\end{algorithm}

Comparing MD-GPS to EM, the C-step is equivalent to the E-step and S-step equivalent to the M-step, with the loss $\ell$ being the negative of the expected discounted sum of returns over trajectories. MD-GPS assumes that the S-step can perfectly minimise the objective in the case of linear dynamics and quadratic costs \cite{li2004iterative}, resulting in exact mirror descent. In practice, most applications of MD-GPS will not satisfy such constraints and so are akin to approximate mirror descent. However, as long as a constraint between the local and global policy can be enforced in terms of KL, various bounds on cost of the global policy can be constructed, see \citet{montgomery2016guided} for further details.

\clearpage
\section{Statistical Precipice}

\label{app:stat-prec}

\subsection{Overview}

Advancements in computational power and algorithmic capabilities have led to a shift in the evaluation of reinforcement learning algorithms, now typically assessed through extensive suites of tasks. Performance metrics such as the mean or median score per task are commonly used, but these metrics can fail to account for the statistical uncertainties arising from a limited number of runs and varying random seeds. The trend towards computationally intensive benchmarks further complicates the issue, as each run can span from hours to weeks, making it impractical to conduct numerous runs per task and thereby increasing the uncertainty in the reported metrics. To address these challenges, we adopt the evaluation methodologies proposed by \citet{agarwal2021deep} and \citet{gorsane2022towards}.

Each algorithm is evaluated across $M$ tasks within a specified environment suite, with $N$ independent runs per task $m \in M$. During each run $n \in N$, performance is measured over $E$ episodes, each consisting of $T$ timesteps. At each interval $i$, the mean return $G^i_{m,n}$ is computed, and the model is checkpointed. The model with the highest mean return across all intervals is selected for final evaluation.

Following the completion of a training run $n \in N$, the best model is further evaluated over $10 \times E$ episodes. We normalise scores $x_{m,n}$ for each task $m = 1, \ldots, M$ and run $n = 1, \ldots, N$, scaling them based on the minimum and maximum scores observed across all runs. This normalization produces a set of normalised scores $x_{1:M,1:N}$ per algorithm. These scores are then aggregated into a single scalar estimate, $\Bar{x}$.

To ensure robust statistical confidence, we employ a 95\% confidence interval derived from stratified bootstrapping over the $M \times N$ experiments, treating these as random samples. This method integrates the performance across all tasks and runs, simulating the statistical reliability of multiple runs on a single task while considering task diversity. Results are reported for the entire suite.

\textbf{Metrics}:

We utilize normalised scores to evaluate algorithm performance, employing metrics that go beyond simple median and mean calculations:
\begin{itemize}
\item \textbf{Interquartile Mean (IQM):} This metric calculates the mean of the central 50\% of runs, excluding the lower and upper 25\%. It is more robust to outliers than the mean and less biased than the median, offering higher statistical efficiency and detecting improvements with fewer runs \cite{agarwal2021deep}.
\item \textbf{Probability of Improvement:} This measures the likelihood that algorithm X will outperform algorithm Y on a random task $m$, using the Mann-Whitney U-statistic. It is defined as:
\begin{equation}
Pr(X>Y) = \frac{1}{M}\sum_{m=1}^{M}Pr(X_m > Y_m)
\end{equation}
    \begin{equation}
    Pr(X_m > Y_m) = \frac{1}{NK}\sum_{i=1}^{N}\sum_{j=1}^{K}S(x_{m,i}, y_{m,j})
    \end{equation}
\begin{equation}
    S(x, y) = \begin{cases}
        1, & \text{if } y<x \\
        \frac{1}{2}, & \text{if } y=x \\
        0, & \text{if } y>x \\
    \end{cases}
\end{equation}
Statistical significance is determined using the Neyman-Pearson criterion, based on the confidence interval bounds \cite{bouthillier2021accounting}.
\end{itemize}
Additionally, performance profiles can be employed to visually compare methods, illustrating the fraction of runs that exceed a given threshold. These profiles aid in identifying stochastic dominance and empirical performance bounds. We also plot the interquartile mean score against environment steps to evaluate sample efficiency.

\subsection{Hyperparameters}

\begin{table}[H]
\centering
\caption{Statistical Precipice Evaluation Hyperparameters}
\label{tab:evaluation_hyperparameters}
\begin{tabular}{@{}ll@{}}
\toprule
Parameter                          & Value \\
\midrule
Number of Games for In-Training Evaluation - $E$ & 128   \\
Number of Games for End-of-Training Evaluation - $10 \times E$ & 1280  \\
Seeds per environment - $N$ & 5     \\
Tasks per suite - $M$ & Continuous=3, Discrete=2\\
\bottomrule
\end{tabular}
\end{table}

\vspace{-1cm}




\end{document}